\documentclass[10pt,twocolumn,letterpaper]{article}

\usepackage{cvpr}              

\definecolor{cvprblue}{rgb}{0.21,0.49,0.74}
\definecolor{textgreen}{rgb}{0.3,0.7,0.3}
\definecolor{textblue}{rgb}{0.2,0.5,0.9}
\definecolor{lightblue}{rgb}{0.75,0.89,0.96}
\definecolor{lightgreen}{rgb}{0.85,0.94,0.85}
\definecolor{lightpeach}{rgb}{1.0,0.9,0.8} 
\definecolor{lightyellow}{rgb}{1.0, 0.98, 0.8} 
\definecolor{lightlavender}{rgb}{0.9,0.85,1.0} 
\definecolor{softcoral}{rgb}{1.0, 0.75, 0.7} 
\definecolor{maroon}{cmyk}{0,0.87,0.68,0.32}
\definecolor{lightgray}{gray}{0.8} 
\usepackage[pagebackref,breaklinks,colorlinks,allcolors=cvprblue]{hyperref}

\usepackage[accsupp]{axessibility}
\usepackage{arydshln} 
\usepackage{multirow}
\usepackage{makecell}
\usepackage{xcolor}
\usepackage{colortbl}
\usepackage{booktabs} 
\usepackage{bm}
\usepackage{mathtools}

\def\paperID{4938} 
\def\confName{CVPR}
\def\confYear{2025}

\title{
Empowering LLMs to Understand and Generate Complex Vector Graphics
}

\author{
Ximing Xing$^{1}$, Juncheng Hu$^{1}$, Guotao Liang$^{1}$, Jing Zhang$^{1}$, Dong Xu$^{2}$, Qian Yu$^{1}$\thanks{Corresponding author.} \\
$^{1}$Beihang University, China \\
{\tt\small \{ximingxing, hujuncheng, liangguotao, zhang\_jing, qianyu\}@buaa.edu.cn} \\
$^{2}$The University of Hong Kong, China \\
{\tt\small dongxu@cs.hku.hk}
}

\begin{document}

\twocolumn[{
\renewcommand\twocolumn[1][]{#1}
\maketitle
\begin{center}
\vspace{-0.5em}
\captionsetup{type=figure}
\includegraphics[width=1.0\textwidth]{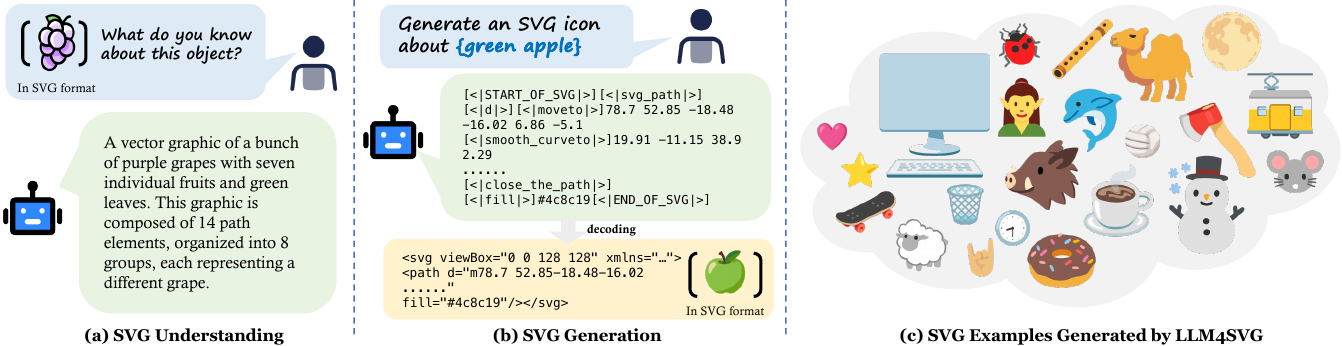}
\vspace{-2em}
\captionof{figure}{
\textbf{Our LLM4SVG can understand and generate vector graphics from textual description.}
Our LLM4SVG is designed to:
\textbf{(a)} Understand the semantics of SVG (Scalable Vector Graphics) source code and directly extract the meanings conveyed by vector images;
\textbf{(b)} Generate corresponding structured SVG representations from textual prompts and decode them into SVG source code that accurately reflects the described content.
\textbf{(c)} illustrates some SVG examples generated by our method.
}
\label{fig:teaser}
\end{center}
}]

\renewcommand{\thefootnote}{\fnsymbol{footnote}}
\footnotetext[1]{Corresponding author.}

\begin{abstract}
The unprecedented advancements in Large Language Models (LLMs) have profoundly impacted natural language processing but have yet to fully embrace the realm of scalable vector graphics (SVG) generation. While LLMs encode partial knowledge of SVG data from web pages during training, recent findings suggest that semantically ambiguous and tokenized representations within LLMs may result in hallucinations in vector primitive predictions. Additionally, LLM training typically lacks modeling and understanding of the rendering sequence of vector paths, which can lead to occlusion between output vector primitives. In this paper, we present \textbf{LLM4SVG}, an initial yet substantial step toward bridging this gap by enabling LLMs to better understand and generate vector graphics. LLM4SVG facilitates a deeper understanding of SVG components through learnable semantic tokens, which precisely encode these tokens and their corresponding properties to generate semantically aligned SVG outputs. Using a series of learnable semantic tokens, a structured dataset for instruction following is developed to support comprehension and generation across two primary tasks. Our method introduces a modular architecture to existing large language models, integrating semantic tags, vector instruction encoders, fine-tuned commands, and powerful LLMs to tightly combine geometric, appearance, and language information. To overcome the scarcity of SVG-text instruction data, we developed an automated data generation pipeline that collected our \textbf{SVGX-SFT Dataset}, consisting of high-quality human-designed SVGs and 580k SVG instruction following data specifically crafted for LLM training, which facilitated the adoption of the supervised fine-tuning strategy popular in LLM development. By exploring various training strategies, we developed LLM4SVG, which significantly moves beyond optimized rendering-based approaches and language-model-based baselines to achieve remarkable results in human evaluation tasks. Code, model, and data will be released at: \href{https://ximinng.github.io/LLM4SVGProject/}{https://ximinng.github.io/LLM4SVGProject/}
\end{abstract}    
\section{Introduction}
\label{sec:intro}
Scalable Vector Graphics (SVGs) constitute a fundamental image encoding paradigm wherein visual elements are constructed from primitive geometric entities defined by mathematical formulations, contrasting with raster graphics' discrete pixel matrices. Vector-based representation offers resolution independence, preserving geometric precision across arbitrary scaling transformations without degradation.
Vector graphics exhibit superior compression efficiency, optimizing storage requirements and transmission bandwidth. Their parametric editability enables precise manipulation of constituent elements—a characteristic instrumental during iterative design processes. These mathematical underpinnings render SVGs exceptionally suitable for applications demanding visual fluency and precision.

In recent years, there has been a significant increase in interest in vector graphics generation~\cite{clipdraw_frans_2022,clipclop_mirowski_2022,Clipasso_vinker_2022,vectorfusion_jain_2023,diffsketcher_xing_2023,svgdreamer_xing_2023,supersvg_hu_2024,starvector_Rodriguez_2023,NIVeL_thamizharasan_2024,T2VecNeualPath_zhang_2024}.
Notwithstanding the significant advantages inherent to SVGs, current deep learning-based generative methods still face limitations in producing high-quality, complex SVG outputs.
The current approaches~\cite{sketchrnn_david_2018,svgvae_lopes_2019,deepsvg_carlier_2020,im2vec_reddy_2021,deepvecfont_wang_2021,iconshop_wu_2023,strokenuwa_tang_2024} represent SVGs using a restricted command path and leverages sequential model learning.
Such methods predominantly engage with simplified SVGs, confined to basic path commands (\textit{e.g.} move to, line to, cubic bézier) and are frequently limited in complexity; certain approaches focus exclusively on fundamental fonts~\cite{svgvae_lopes_2019} or icons~\cite{deepvecfont_wang_2021}.
Recent innovations~\cite{vectorfusion_jain_2023,diffsketcher_xing_2023,svgdreamer_xing_2023,NIVeL_thamizharasan_2024,T2VecNeualPath_zhang_2024} have incorporated advanced image diffusion models~\cite{ldm_Rombach_2022,dreamfusion_poole_2023} to facilitate the generation of raster images, subsequently translated into SVG format via a differentiable rasterizer~\cite{diffvg_Li_2020} predicated on bézier curve representations.
While the utilization of generative raster images introduces a degree of variety, this process is characterized by an cumbersome iterative procedure, and the resultant SVGs remain non-editable and fail to align with the expectations of professional designers.
In light of these developments, a critical gap persists in the realm of systems capable of directly synthesizing intricate and detailed SVG code, fully leveraging the comprehensive array of SVG primitives requisite for sophisticated design applications.

Recent advancements in large language models (LLMs)~\cite{ChatGPT,GPT4,claude3.5,LLaMA_Touvron_2023,llava_Liu_2023} have evidenced their capacity to comprehend and parse XML syntax~\cite{Copilot_chen_2021}, achieved through extensive pre-training on a diverse corpus of text data sourced from the Web. 
This proficiency establishes a robust foundation for LLMs to synthesize vector graphics~\cite{starvector_Rodriguez_2023}. 
Notably, state-of-the-art models, such as GPT-4~\cite{GPT4} and Claude~\cite{claude3.5}, show proficiency in generating simple vector primitives like triangles and rectangles; however, they often encounter significant limitations in synthesizing complex graphics.
This is because, during pre-training, SVG data from the internet is often embedded within web page code, requiring LLMs to parse layered languages, which renders SVG data less accessible amidst lengthy XML tags.
These limitations manifest as confusion or hallucinations in vector path sequences, yielding semantically ambiguous graphics and improperly encoded attributes.

In this paper, we present LLM4SVG, an initial yet substantial step toward bridging this gap by enabling LLMs to better understand and generate vector graphics, as illustrated in Fig.~\ref{fig:teaser}.
Built on an existing LLM/MLLM, LLM4SVG maximizes the model's potential for vector graphic synthesis.
Our paper makes the following contributions:
\begin{itemize}
\item \textbf{LLM4SVG: The First Framework Supporting Arbitrary LLMs for SVG Tasks.} We present LLM4SVG, a framework enabling any LLM or MLLM to understand and generate SVGs effectively. Our approach addresses vector primitive prediction challenges through learnable semantic tokens, precisely encoding SVG components and properties. This bridges natural language processing and vector graphics, aligning outputs with human design principles.
\item \textbf{Modular Architecture with Decoupled Vector Instructions and Parameters.} LLM4SVG enhances traditional LLM architectures by decoupling vector instructions from parameters. This integration enables comprehensive understanding of SVG elements, producing semantically consistent vector graphics by merging geometry, appearance, and linguistic information.
\item \textbf{Development of SVGX-SFT Dataset.} We introduce the SVGX-SFT Dataset with high-quality human-designed SVGs and 580k instruction following data crafted for LLM training. This dataset supports our training strategy, enhancing model performance in SVG generation and establishing a foundation for future vector graphics research.
\end{itemize}

\section{Related Work}
\label{sec:related_work}

\subsection{Vector Graphics Generation}
Scalable Vector Graphics (SVGs) provide a declarative format for visual concepts articulated through primitives.
One approach to generating SVG content entails training a neural network to generate predefined SVG commands and attributes~\cite{sketchrnn_david_2018,svgvae_lopes_2019,deepsvg_carlier_2020,im2vec_reddy_2021,deepvecfont_wang_2021,iconshop_wu_2023,strokenuwa_tang_2024}. 
Neural networks designed for learning SVG representations typically include architectures such as RNNs~\cite{sketchrnn_david_2018,im2vec_reddy_2021}, VAEs~\cite{svgvae_lopes_2019,deepsvg_carlier_2020,strokenuwa_tang_2024}, and Transformers~\cite{deepsvg_carlier_2020,deepvecfont_wang_2021,iconshop_wu_2023}. The training of these networks is heavily dependent on datasets in vector form.
However, the limited availability of large-scale vector datasets significantly constrains their generalization capability and their ability to synthesize intricate vector graphics.

Li \textit{et al.}~\cite{diffvg_Li_2020} introduce a differentiable rasterizer that bridges the vector graphics and raster image domains. While image generation methods that traditionally operate over vector graphics require a vector-based dataset, this approach for SVG generation~\cite{evolution_tian_2022, LIVE_Ma_2022, marvel_su_2023,ClipGen_Shen_2022, CLIPVG_song_2023,supersvg_hu_2024} involves directly optimizing the geometric and color parameters of SVG paths using the guidance of a pretrained vision-language model.
Recent advances in visual text embedding contrastive language-image pre-training model (CLIP)~\cite{CLIP_radford_2021} has enabled a number of successful methods~\cite{clipdraw_frans_2022, clipclop_mirowski_2022, Clipasso_vinker_2022} for synthesizing sketches. 
In contrast to CLIP, several methods~\cite{vectorfusion_jain_2023,diffsketcher_xing_2023,svgdreamer_xing_2023} integrate diffusion models with differentiable rasterizers to achieve superior generation capabilities and enhanced image consistency.
Moveover, recent studies~\cite{NIVeL_thamizharasan_2024, T2VecNeualPath_zhang_2024} combine optimization-based approaches with neural networks to learn vector representations, incorporating geometric constraints into vector graphics.

\subsection{Vector Graphics Understanding}
Recent research in vector graphics has advanced both recognition and evaluation methodologies. YOLAT~\cite{yolat_jiang_2021} pioneered treating vector graphics recognition as a detection problem without rasterization, though it struggles with complex hierarchical structures and semantic relationships. YOLAT++~\cite{yolat++_dou_2024} addressed these limitations by introducing hierarchical recognition capabilities and a new chart-based dataset. 
Meanwhile, VGBench~\cite{vgbench_zou_2024} developed a comprehensive benchmark evaluating LLMs on vector graphics understanding and generation across various formats, revealing that while LLMs show promise, they perform poorly on low-level formats like SVG. 
Despite these advances, existing approaches remain constrained—recognition models lack generation capabilities, evaluation frameworks don't address fundamental representation issues, and none adequately capture the bidirectional relationship between natural language and vector graphics necessary for creative design workflows. 

\subsection{Large Language Models}
Large Language Models (LLMs) have made significant strides in natural language understanding, demonstrating strong generalization and reasoning abilities through extensive pre-training on large-scale text corpora~\cite{ChatGPT,GPT4,LLaMA_Touvron_2023,claude3.5,qwen2.5,geminipro,Grok2,yi_young_2024,falcon_almazrouei_2023,phi2,phi2_javaheripi_2023,phi3_abdin_2024}.

LLMs can be broadly categorized into two types. The first type serves as an interface for individual modality-specific models~\cite{vispro_Gupta_2023,audiogpt_huang_2023,gorilla_patil_2023}, eliminating the need for retraining but relying heavily on external model availability. The second type employs an end-to-end training approach, which can either train models from scratch using large-scale multi-modal datasets~\cite{kosmos_2023_huang,GPT4} or fine-tune pre-trained LLMs for specific applications~\cite{llava_Liu_2023}. Our work follows the latter strategy, adapting pre-trained LLMs to generate and understand vector graphics while maintaining flexibility for multi-modal extensions.

LLMs naturally extend into multi-modal domains, enabling them to process and generate content beyond text. Multi-modal Large Language Models (MLLMs) leverage this capability to handle diverse modalities, including images~\cite{multimodalgpt_gong_2023,kosmos_2023_huang,llava_Liu_2023,visionllm_wang_2023,minigpt4_zhu_2024,qwen2_wang_2024}, audio~\cite{pengi_deshmukh_2023,audiogpt_huang_2023}, motion~\cite{motiongpt_jiang_2023,videollama_zhang_2023}, and 3D point clouds~\cite{3dllm_hong_2023,pointllm_xu_2024}. This adaptability makes LLMs a powerful foundation for expanding into vector graphics and other specialized tasks.

\subsection{Instruction Tuning}
In natural language processing, researchers have explored various methods~\cite{training_ouyang_2022, Self-instruct_wang_2022, ChatGPT} for instruction-tuning LLMs to improve their ability to follow natural language instructions and perform real-world tasks. This straightforward approach has been shown to significantly enhance LLMs' zero-shot and few-shot generalization capabilities. With the rise of multimodal large language models, LLaVA~\cite{llava_Liu_2023} has leveraged visual instruction tuning with vision-language data, greatly improving open-source MLLMs' performance on multimodal tasks.

Recently, studies have investigated fine-tuning LLMs with image embeddings to generate Scalable Vector Graphics (SVG) by treating SVG code as a text-based representation~\cite{starvector_Rodriguez_2023, exploring_xu_2024}. While promising, these approaches often overlook the hierarchical and structured nature of SVG files, treating them merely as sequential text.

\section{SVGX-SFT Dataset}
\label{sec:instruction_data}
\newcommand{\prompt}[1]{\textcolor{textblue}{#1}}
\newcommand{\img}[1]{\colorbox{lightpeach}{#1}}
\newcommand{\elem}[1]{\colorbox{lightgreen}{#1}}
\newcommand{\attrval}[1]{\colorbox{lightyellow}{#1}}
\newcommand{\svgpath}[1]{\colorbox{softcoral}{#1}}

\begin{table*}[t]
\centering
\begin{minipage}{0.52\textwidth}
    \centering
    \resizebox{\linewidth}{!}{
    \begin{tabular}{l|c|l} 
    \toprule
    \textbf{Type} & \textbf{Role} & \multicolumn{1}{c}{\textbf{Content Template}} \\
    \midrule
    \multirow{3}{*}{\# 1} & SYSTEM & You are a helpful assistant, please help me generate SVG \texttt{</s>} \\
    & USER & Generate an SVG illustration from the given description: \prompt{$\{\mathrm{prompt}\}$} \texttt{</s>} \\
    & ASSISTANT & \elem{SOV} 
    \elem{Path} \elem{MoveTo} \attrval{Coord} \elem{LineTo} \attrval{Coord} \dots \elem{FILL} \attrval{RGB} \dots \elem{EOV} \texttt{</s>} \\
    \midrule
    \multirow{3}{*}{\# 2} & SYSTEM & You are a helpful assistant, please help me generate an SVG from this image and description. \texttt{</s>} \\
    & USER & Refer to rendering image: \prompt{$\{\mathrm{img}\}$} and generate SVG from the given description: \prompt{$\{\mathrm{prompt}\}$} \texttt{</s>} \\
    & ASSISTANT & \elem{SOV} 
    \elem{Path} \elem{MoveTo} \attrval{Coord} \elem{LineTo} \attrval{Coord} \dots \elem{FILL} \attrval{RGB} \dots \elem{EOV} \texttt{</s>} \\
    \midrule
    \multirow{3}{*}{\# 3} & SYSTEM & Attempt to identify this SVG \texttt{</s>} \\
    & USER & \makecell[l]{
    The following is an SVG illustration: \elem{SOV} \svgpath{$\text{Path}_1$} \dots \svgpath{$\text{Path}_n$} \elem{EOV}
    \\
    } \\
    & ASSISTANT & \makecell[l]{
    Text description of this SVG: \prompt{$\{\mathrm{desc}\}$}
    \texttt{</s>}
    } 
    \\
    \midrule
    \multirow{3}{*}{\# 4} & SYSTEM & Describe this SVG based on its image representation \texttt{</s>} \\
    & USER & \makecell[l]{
        The following is an SVG illustration: \elem{SOV} \svgpath{$\text{Path}_1$} \dots \svgpath{$\text{Path}_n$} \elem{EOV} 
        rendering result: \prompt{$\{\mathrm{img}\}$}
    } \\
    & ASSISTANT & \makecell[l]{
        Text description of this SVG: \prompt{$\{\mathrm{desc}\}$}.
        This SVG contains \prompt{$\{{\textrm{n\_path}}\}$} primitives.
        \texttt{</s>}
    }  \\
    \midrule
    \multirow{6}{*}{\# 5} & SYSTEM & Describe this SVG based on its image representation \texttt{</s>} \\
    & USER & \makecell[l]{
        SVG group 1: \elem{GROUP} \svgpath{$\text{Path}_1$} \dots \svgpath{$\text{Path}_3$} \elem{EOG} 
        rendering result: \prompt{$\{\mathrm{img}\}_{1}$}
        \\
        SVG group 2: \elem{GROUP} \svgpath{$\text{Path}_6$} \dots \svgpath{$\text{Path}_9$} \elem{EOG} 
        rendering result: \prompt{$\{\mathrm{img}\}_{2}$} \texttt{</s>}
    } \\
    & ASSISTANT & \makecell[l]{
        Text description of this SVG: \prompt{$\{\mathrm{desc}\}$}
        \\
        The 1st SVG group contains \prompt{$\{{\textrm{n\_paths}}\}_1$} primitives representing \prompt{$\{\mathrm{desc}\}_{1}$}
        \\
        The 2nd SVG group contains \prompt{$\{{\textrm{n\_paths}}\}_2$} primitives representing \prompt{$\{\mathrm{desc}\}_{2}$}
        \texttt{</s>}
    }  \\
    \bottomrule
    \end{tabular}
    }
    \caption{
    \textbf{Instruction Following Template.}
    We developed five distinct instruction templates tailored for tasks in vector graphics generation and understanding. 
    Specifically, Types \#1 and \#2 facilitate the generation task, while Types \#3, \#4, and \#5 focus on the understanding task.
    \textcolor{textblue}{\{prompt\}} denotes a brief image caption generated via BLIP~\cite{blip_li_2022}, \prompt{$\{{\textrm{n\_paths}}\}_i$} represents the total number of primitives in group $i$, \textcolor{textblue}{\{desc\}} provides a detailed GPT-4~\cite{GPT4} generated description, and \elem{Token} represents different types of SVG semantic tokens.
    \svgpath{$\text{Path}_i$} serves as the representation of a complete SVG primitive, encompassing a structured set of SVG semantic tokens essential for comprehensive vector graphic description.
    Types \#1$\sim$\#5 provide a structured framework for training SVG semantic tokens, facilitating more accurate vector representation and understanding.
    Losses are computed only on model responses.
    \texttt{</s>} indicates the end-of-sentence token. 
    ``SYSTEM'' is an instruction that describes the type of task, specifically the context of the conversation. ``ASSISTANT'' denotes the output generated in response to the instruction, representing the LLM's reply. ``USER'' refers to the input data provided by the user.
    }
    \label{tab:instruct_template}
\end{minipage}%
\hfill
\begin{minipage}{0.46\textwidth}
    \centering
    \includegraphics[width=\linewidth]{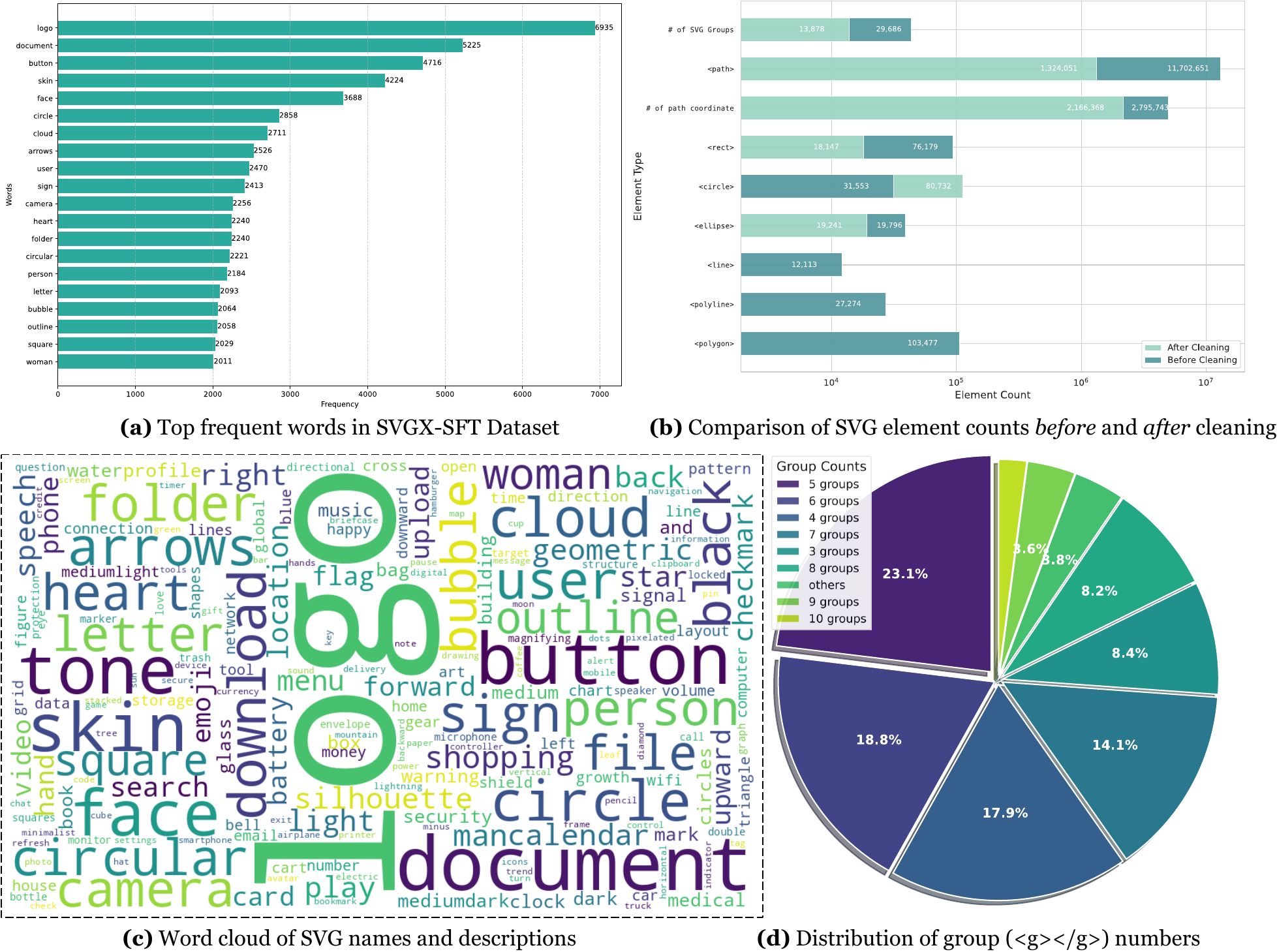}
    \captionof{figure}{
    \textbf{Overview of SVGX-SFT Dataset.}
    \textbf{(a)} \textit{Top Frequent Words in SVGX-SFT Dataset.}
    The most frequently occurring words in the SVGX-SFT dataset, highlighting common patterns and terminology used in SVG metadata.
    \textbf{(b)} \textit{Comparison of SVG Element Counts Before and After Cleaning.}
    A comparison of the number of SVG elements before and after the cleaning process. The reduction in element count demonstrates the effectiveness of our preprocessing steps.
    \textbf{(c)} \textit{Word Cloud of SVG Names and Descriptions.}
    A word cloud visualization of SVG names and descriptions, illustrating the distribution and emphasis of different terms in the dataset.
    \textbf{(d)} \textit{Group (\texttt{<g></g>}) Number Distribution.}
    The distribution of \texttt{<g>} (group) elements in the dataset, showing the frequency of grouped elements and their structural significance within SVG files.
    }
    \label{fig:sft_analysis}
\end{minipage}
\vspace{-1.5em}
\end{table*}
A significant obstacle in developing an end-to-end LLM is the acquisition of large-scale instruction-following data, which is indispensable for representation learning, aligning latent spaces, and guiding models to align with human intent~\cite{llava_Liu_2023}. In the domain of vector graphics, this obstacle is particularly acute, as the high production cost and tagging challenges associated with vector graphics constrain current research to a limited scope of applications.
These areas include simple human hand drawings~\cite{sketchshoe_yu_2016,sketchy_Sangkloy_2016}, fonts~\cite{svgvae_lopes_2019}, and iconic graphics~\cite{figr8_clouatre_2019,deepsvg_carlier_2020}.

To address these challenges, we manually collected approximately $250,000$ colorful and complex vector graphics and developed a normalization process to ensure that the collected data conformed to a consistent standard, including uniform canvas size, relative coordinate systems, and representation.
These high-quality vector datasets provided us with a solid foundation for the development of LLM4SVG.
Additionally, inspired by the recent success of GPT models in text annotation tasks~\cite{chatgptannotation_gilardi_2023}, we utilized BLIP~\cite{blip_li_2022} to annotate rasterized vector graphs and GPT-4~\cite{GPT4} for instruction-following data collection.

\noindent\textbf{SVG Re-captioning.} 
SVG data collected from the Internet often contains noise, and directly using it for learning can compromise the model's potential representational accuracy.
Approximately half of the data in an SVG file is redundant for visual rendering.
This redundancy includes: (1) temporary data used by vector editing applications, (2) non-optimal structural representations of SVG, and (3) unused and invisible graphic elements.
We propose an SVG data preprocessing pipeline designed to losslessly reduce the size of SVG files generated by vector editing applications. 
Details are provided in Sec.~\ref{sec:supp_datasets} and Fig.~\ref{fig:supp_clean} of Supplementary.
After optimizing the SVG, we rasterize it into an image of $512\times512$ pixels and use the BLIP~\cite{blip_li_2022} model to generate a corresponding caption as a text prompt.
Consequently, we obtain a multimodal dataset, each entry of which is a triplet consisting of the optimized SVG, its corresponding rasterized image, and a text description generated by the BLIP model.

Subsequently, we propose a strategy for the automatic generation of SVG instruction-following data. As outlined in Table~\ref{tab:instruct_template}, the instruction data are categorized into two distinct parts: the first (items \#1 and \#2) addresses the synthesis of vector graphics, while the second part (items \#3, \#4, \#5) pertains to the comprehension of vector graphics.

\begin{figure*}[t]
\centering
\includegraphics[width=1.0\linewidth]{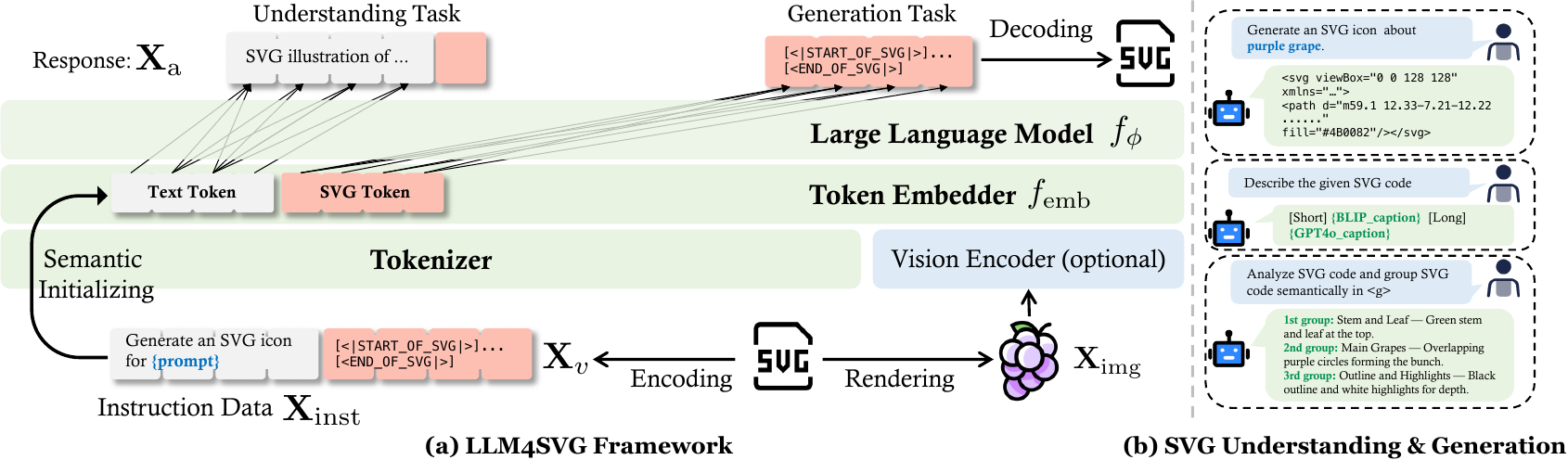}
\vspace{-1.5em}
\caption{
\textbf{An Overview of LLM4SVG.} 
Our LLM4SVG is capable of understanding and generating SVGs effectively. \textbf{(1)} During the training phase, we provide both the original SVG code $\mathbf{X}_{v}$ and the corresponding instruction data $\mathbf{X}_{\mathrm{inst}}$ as input. For the understanding task, we use detailed descriptions $\mathbf{X}_{a}$ generated by GPT-4~\cite{GPT4} as the training labels. For the generation task, the SVG code portion is masked and serves as the target that the model needs to predict. \textbf{(2)} During the inference phase, for the understanding task, given an SVG source code, the model generates a description that aligns with the semantics expressed by the SVG. For the generation task, the model generates an SVG based on the input text prompt. During both training and inference phases, the rendered image $\mathbf{X}_{\mathrm{img}}$ of the SVG can be used as conditional input to the model, guiding the content that the model understands or generates.
} \label{fig:pipeline}
\vspace{-1.5em}
\end{figure*}

\noindent\textbf{SVG Instruction Following Data.}
The constructed dataset adheres to a standardized instruction format, as depicted in Table~\ref{tab:instruct_template} \#1, comprising Text-SVG pairs for fine-tuning in text-to-SVG generation tasks.
For text-guided SVG synthesis, visual prompts are indispensable. As illustrated in Table~\ref{tab:instruct_template} \#2, Text-Image-SVG triples facilitate instruction tuning for text-and-image to SVG generation.

During the SVG re-captioning phase, we obtained the corresponding text descriptions from the BLIP based on the rendering results, which were sufficient for text prompts, but too short for comments to understand SVG.
Inspired by the recent success of GPT models in text annotation tasks~\cite{chatgptannotation_gilardi_2023}, we utilized ChatGPT~\cite{ChatGPT}/GPT-4~\cite{GPT4} for instruction-following data generation.

In total, we collected a dataset consisting of 250k annotated, high-quality, and standardized vector graphics, along with 580k unique SVG-Text-Image samples. The distribution across sample types is as follows: 250k samples of type \#1, which extend to 250k samples of type \#2, 60k samples of types \#3 and \#4, and 20k samples of type \#5.

\section{Instruction Tuning Scheme}
\label{sec:model}
We then delve into the architecture of LLM4SVG, which takes as input an SVG and user instruction and outputs responses. 
We first introduce the definition of a semantic token, and then introduce the two-stage training strategy.

\subsection{SVG Semantic Tokens}
\label{sec:semantic_token}
For an input SVG $\mathbf{X}_v$, we convert it from raw code into a structured representation.
To accomplish this, we defined 55 SVG semantic tokens $s_i \in \mathbb{R}^{55}$ (including 15 tag tokens, 30 attribute tokens, 10 path command tokens, details are provided in Sec.~\ref{sec:supp_svg_tokens} and Fig.~\ref{tab:supp_svg_tokens} of Supplementary).
These SVG tokens are used to replace all tags and attributes in the SVG source code, thus preventing the textual encoding of SVG tags and attributes as regular text. 
For example, the tag \texttt{<path>} will be tokenized as an SVG semantic token, rather than the literal ``path'' by the tokenizer.
This ensures the uniqueness of SVG tags and attributes, and allows for their efficient integration into LLMs in a manner that is consistent with SVG definitions and optimizes token initialization.
We adapt the embedding layer $\mathbf{W}_{\mathrm{emb}} \in \mathbb{R}^{|\mathcal{V}|'}$ to learn the embeddings of these new tokens, where $|\mathcal{V}|' := |\mathcal{V}| + 55$ represents the sum of the size of the original vocabulary and the additional SVG tokens. Each new token is initialized based on the semantic average of its descriptive text $s$, as defined by the equation:
\begin{equation}
\bm{E}(s) = \frac{1}{n}\sum_{j=1}^n \mathbf{W}_{\mathrm{emb}}^{\top} \cdot w_{j}
\label{eq:svg_token_init}
\end{equation}
\noindent where $w_{j}$ represents the $j$-th description token of $s$, $\bm{E}(\cdot)$ represents the token embedding layer and $n$ represents the length of the token after encoding the description $s$. This initialization provides a good starting point for each SVG token and builds a compact, distributed representation for all SVG tokens.

\subsection{Architecture}
\label{sec:arch}
The primary goal is to effectively leverage the capabilities of both the pre-trained LLMs and visual models. The network architecture is illustrated in Fig.~\ref{fig:pipeline}. 
We chose GPT-2~\cite{gpt2_radford_2019}, Phi-2~\cite{phi2,phi2_javaheripi_2023} and Falcon~\cite{falcon_almazrouei_2023} as the foundational LLMs for our framework, denoted as $f_{\phi}$ and parameterized by $\phi$. 
These models were chosen because they possess the ability to understand both visual and textual data, and they demonstrate effective instruction-following properties in various language tasks among existing open source models.
Theoretically, other LLMs with similar capabilities could also serve as the bases for our method.

\subsection{Training}
For each SVG $\mathbf{X}_\mathrm{v}$, we sample multi-turn conversation data ($\mathbf{X}_{\textrm{un}}^1$, $\mathbf{X}_{\textrm{gen}}^1$, \dots, $\mathbf{X}_{\textrm{un}}^T$, $\mathbf{X}_{\textrm{gen}}^T$) from our SVG instruction dataset, where $\mathbf{X}_{\textrm{un}}$ represents the understanding tasks and $\mathbf{X}_{\textrm{gen}}$ refers to the generation tasks. $T$ denotes the total number of turns in the conversation.

We apply instruction-tuning to the LLM using its original auto-regressive training objective for enhancing its performance on prediction tasks.
Specifically, for a sequence of length $L$, we compute the probability of the target answers $\mathbf{X}_\mathrm{a}$ using the following equation:
\begin{equation}
p(\mathbf{X}_{\mathrm{a}} | \mathbf{X}_{\mathrm{v}}, \mathbf{X}_{\mathrm{inst}})
= \prod_{i=1}^L p_{\theta} ( \bm{x}_i | \mathbf{X}_{\mathrm{v}}, \mathbf{X}_{\mathrm{inst}},\mathbf{X}_{\mathrm{a}}, \hat{\bm{x}}_{i-1})
\label{eq:x_pred_likehood}
\end{equation}
\noindent where $\theta$ represents the trainable parameters of the model, $\mathbf{X}_{\mathrm{inst}}$ and $\mathbf{X}_{\mathrm{a}}$ are the tokens corresponding to the instructions and answers for all preceding turns before the current prediction token $\bm{x}_i$, respectively. The term $\hat{\bm{x}}_{i-1} \coloneqq (\bm{x}_{i-1}, \cdots, \bm{x}_{1})$ denotes the sequence of tokens that have been predicted in previous steps.

\noindent\textbf{Stage 1: Pre-training for Feature Alignment.}
These pairs are converted to instruction-following data using the naive expansion method describe in Sec.~\ref{sec:instruction_data}. 
Each sample can be treated as a single-turn conversation. 
To construct the input $\mathbf{X}_{\mathrm{inst}}$, instruction data is randomly sampled. 
In training, we keep the weights of both the visual encoder and the LLM frozen, and focus on maximizing the likelihood of Eq.~\ref{eq:x_pred_likehood} with trainable parameters $\theta = \mathbf{W}_{\mathrm{emb}}$ only, where $\mathbf{W}_{\mathrm{emb}}$ represents the word embedding.

\noindent\textbf{Stage 2: Fine-tuning End-to-End.}
We employ the entire instruction dataset for supervised fine-tuning of all parameters, including those of the LLM, \textit{i.e.}, the trainable parameters are $\theta = \{ \mathbf{W}_{\mathrm{emb}}, \phi \}$ in Eq.~\ref{eq:x_pred_likehood}. 
We consider two specific use case scenarios:
\begin{itemize}
\item \textit{Efficient parameters fine-tuning.}
Methods like LoRA~\cite{lora_hu_2021,qlora_dettmers_2024} only fine-tune a small number of additional model parameters, significantly decreasing computational and storage costs while delivering performance comparable to that of a fully fine-tuned model. This approach facilitates the training and storage of LLMs on consumer hardware.
\item \textit{Full fine-tuning.}
Fine-tuning all parameters requires higher computational resources, especially in large language models.
\end{itemize}

\section{Experiments}
\label{sec:experiments}
\noindent\textbf{Overview.}
This section outlines the core aspects of our model implementation, followed by a description of the dataset collection and preprocessing pipeline, which involves 250,000 high-quality SVG files, as detailed in~\cref{subsec:dataset_analysis}.
We then empirically evaluate the effectiveness of our model, LLM4SVG, in generating high-quality SVGs. This evaluation benchmarks our model against current state-of-the-art methods both qualitatively and quantitatively, as detailed in Sec.~\ref{sec:qualitative} and ~\ref{sec:quantitative}.
This evaluation is augmented with a comprehensive architectural analysis, including ablation studies detailed in Sec.~\ref{subsec:ablation}, to identify the specific contributions of individual model components to the overall performance.

\begin{figure*}
\centering
\includegraphics[width=1\linewidth]{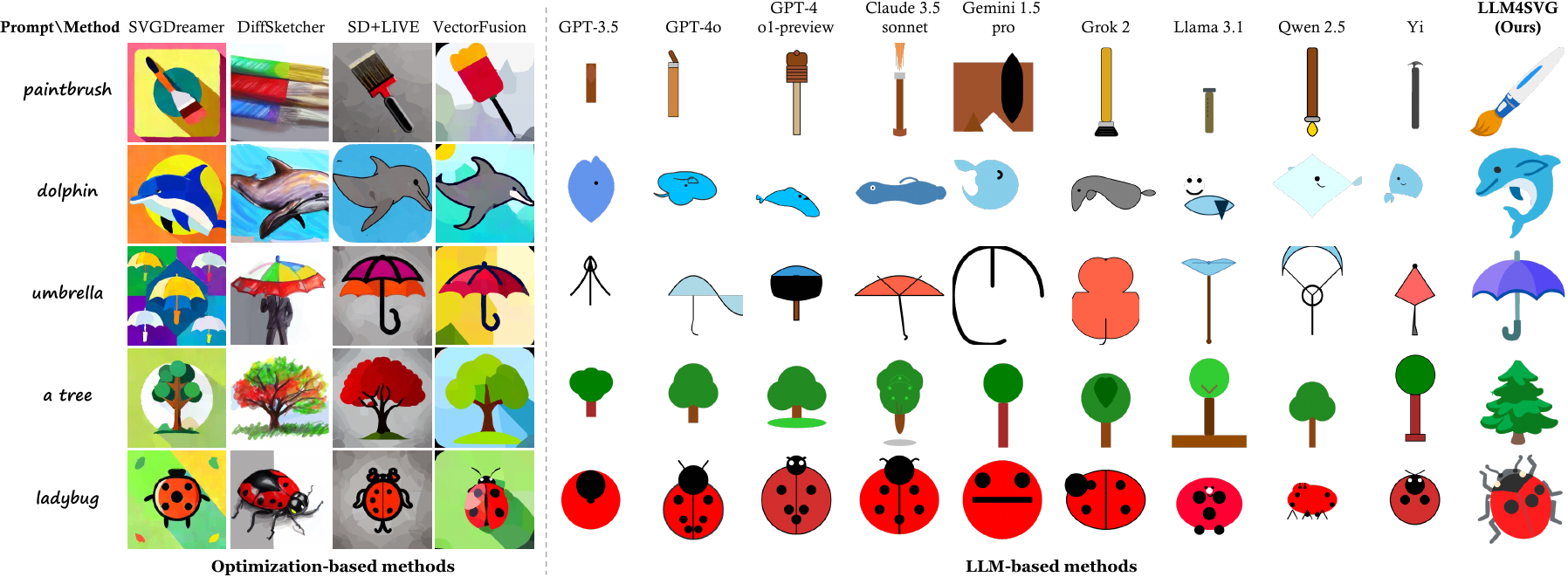}
\vspace{-2em}
\caption{
 \textbf{Qualitative comparison of LLM4SVG with state-of-the-art SVG generation methods}, categorized into optimization-based and LLM-based approaches. The prompts (left) guide the generation of vector graphics across different methods. Optimization-based methods (left section) focus on refining SVGs through iterative optimization, while LLM-based methods (right section) generate SVGs directly from text descriptions. The results highlight differences in abstraction, structure, and fidelity to the input prompt, with our LLM4SVG achieving more structured and visually coherent outputs.
}
\label{fig:compare}
\vspace{-0.5em}
\end{figure*}
\begin{table*}
\centering
\begin{subtable}{0.52\linewidth}
    \centering
    \resizebox{1.0\linewidth}{!}{
    \begin{tabular}{l|c|cccc|c}
    \toprule
    \multirow{2}{*}{Method / Metric}&\multirow{2}{*}{Type}&\multicolumn{4}{c|}{Visual Metric}&Latency\\ 
    \cmidrule{3-7}
&&FID$\downarrow$&CLIPScore$\uparrow$&Aesthetic$\uparrow$&HPS$\uparrow$&Gen. Time$\downarrow$\\
    \midrule
    CLIPDraw~\cite{clipdraw_frans_2022}&Optim-based&
    132.75&0.2486&3.9803&0.2347&5min20s\\
    Evolution~\cite{evolution_tian_2022}&Optim-based&
    123.97&0.1932&4.0845&0.1955&49min42s\\
    $\text{DiffSketcher}$\cite{diffsketcher_xing_2023}&Optim-based&
    77.35&0.2402&4.1562&0.2423&12min9s\\
    LIVE+VectorFusion~\cite{vectorfusion_jain_2023}&Optim-based&
    84.71&0.2298&4.5165&0.2334&32min19s\\
    VectorFusion~\cite{vectorfusion_jain_2023}&Optim-based&
    87.73&0.2720&4.9845&0.2450&11min27s\\
    SVGDreamer~\cite{svgdreamer_xing_2023}&Optim-based&
    72.68&0.3001&5.5432&0.2685&43min56s\\
    \midrule
    SVG-VAE~\cite{svgvae_lopes_2019}&NN-based&
    79.25&0.1893&2.674&0.098&1min4s\\
    DeepSVG~\cite{deepsvg_carlier_2020}&NN-based&
    71.37&0.2118&3.0017&0.109&2min3s\\
    Iconshop~\cite{iconshop_wu_2023}&NN-based&
    85.45&0.2489&3.4682&0.1376&1min8s\\
    StrokeNUWA~\cite{strokenuwa_tang_2024}&NN-based&
    92.31&0.3001&5.5432&0.1659&20s\\
    \midrule
    \textbf{LLM4SVG(GPT-2 small)}&LLM-based&
    78.10&0.3129&5.7327&0.2076&12s\\
    \textbf{LLM4SVG(GPT-2 large)}&LLM-based&
    66.09&0.3205&5.8729&0.2190&14s\\
    \textbf{LLM4SVG(GPT-2-XL)}&LLM-based&
    64.11&0.3496&5.9836&0.2485&18s\\
    \textbf{LLM4SVG(Phi-2)}&LLM-based&
    65.98&0.3373&5.9124&0.2289&20s\\
    \textbf{LLM4SVG(Falcon)}&LLM-based&
    77.13&0.3018&4.9846&0.2012&25s\\
    \textbf{LLM4SVG(LLaVA)}&LLM-based&
    66.72&0.3296&5.6846&0.2177&25s\\
    \bottomrule
    \end{tabular}
    }
   \caption{
    \textbf{Quantitative Comparison between LLM4SVG and State-of-the-Art Text-to-SVG Methods}.
    }
    \label{tab:compare_t2v}
    \end{subtable}
\hfill
\begin{subtable}{0.46\linewidth}
    \centering
    \resizebox{1.0\linewidth}{!}{
    \begin{tabular}{l|c|c|c|c|c|c}
    \toprule
Model&Input&FID$\downarrow$&CLIPScore$\uparrow$&Aesthetic$\uparrow$&HPS$\uparrow$&Avg.Tok\\
    \midrule
    Llama-3.1 70B~\cite{llama3_dubey_2024} &Text&138.44&0.2735&4.3048&0.1665&707.67\\
    \midrule
    Gemini-1.5 Pro~\cite{geminipro} &Text\&Img&145.76&0.2622&4.2708&0.1511&547.50\\
    \midrule
    Claude-3.5~\cite{claude3.5} &Text&82.89&0.3083&5.2370&0.1912&736.38\\
    \midrule
    Yi-1.5 34B~\cite{yi_young_2024} &Text&140.83&0.2824&4.5118&0.1676&633.42\\
    \midrule
    Grok-2~\cite{Grok2} &Text&116.99&0.2840&4.8086&0.1663&581.88\\
    \midrule
    Qwen2.5 70B~\cite{qwen2.5} &Text&131.46&0.2803&4.5024&0.1691&705.00\\
    \midrule
    GPT-3.5~\cite{ChatGPT} &Text\&Img&129.40&0.2949&4.3070&0.1717&530.59\\
    \midrule
    GPT-4o~\cite{GPT4} &Text\&Img&127.78&0.2949&5.0262&0.1788&654.12\\
    \midrule
    GPT-4 o1-preview~\cite{GPT4} &Text\&Img&135.33&0.2968&4.7754&0.1840&755.71\\
    \midrule
    \textbf{LLM4SVG(GPT-2 XL)} &Text&64.11&0.3496&5.9836&0.2485&2297.75\\
    \midrule
    \textbf{LLM4SVG(Falcon)} &Text&77.13&0.3018&4.9846&0.2012&1829.49\\
    \midrule
    \textbf{LLM4SVG(LLaVA)} &Text\&Img&66.72&0.3296&5.6846&0.2177&2009.41\\
    \bottomrule
    \end{tabular}
    }
    \caption{
    \textbf{Quantitative Comparison between LLM4SVG and State-of-the-Art LLMs}.
    }
    \vspace{1em}
\label{tab:compare_llm}
\end{subtable}
\caption{
  \textbf{Quantitative Comparison of LLM4SVG.} (a) is used to indicate comparisons against SVG generation methods. (b) is employed to compare performance with LLM-based methods.
} \label{fig:Quantitative}
\vspace{-1em}
\end{table*}
\noindent\textbf{Implementation Details.}
Our training process consists of two steps. In the first step, we add 55 SVG semantic tokens (Tab.~\ref{tab:supp_svg_tokens}) to the tokenizer and initialize the word embeddings $\mathbf{W}_\mathrm{emb}$ using the semantic average of the description text as described in Eq.~\ref{eq:svg_token_init}.
In the second step, we perform supervised fine-tuning (SFT) using two alternative methods: either LoRA/QLoRA~\cite{lora_hu_2021,qlora_dettmers_2024} fine-tuning or full-parameter fine-tuning. The trainable parameters include the word embeddings and the parameters involved in SFT training. SFT training typically requires 1 to 3 epochs.

During the training process, we apply the AdamW optimizer with hyper-parameters $\beta_1 = 0.9, \beta_2 = 0.999, \epsilon=1\times10^{-8}$. We use a learning rate of $lr=3\times10^{-4}$ with a cosine scheduler type and warmup ratio of 0.1. The training runs for 2 epochs with a maximum token length of 4096. If an SVG's token sequence exceeds this length, it is directly truncated to the maximum length to ensure computational efficiency.
Our implementation is based on LlamaFactory framework~\cite{llamafactory_zheng_2024}. Considering VRAM limitations, we also integrated Unsloth~\cite{unsloth_github_2023} to support efficient training of quantized models. The training was performed using 8 NVIDIA A800 GPUs.

\noindent\textbf{Evaluation Metrics.}
To facilitate a comprehensive assessment of our proposed method compared to baseline approaches, we classify all methods into three categories: Optimization-based methods, Neural Network-based methods, and LLM-based methods. We then evaluate these methods across two key dimensions: visual quality and computational cost. For visual quality, we measure (1) visual quality using FID (Fr\'echet Inception Distance)~\cite{FID_Heusel_2017};
(2) text prompt alignment through the CLIP score~\cite{CLIP_radford_2021}; and (3) aesthetic appeal using the Aesthetic score~\cite{aesthetic_christoph_2022} and HPS (Human Preference Scores)~\cite{HPS_Wu_2023}. 
(4) for computational cost, we test and compare the time cost of generating 10 SVGs by each method.
(5) Avg.Tok represents the length of SVG code after removing comments and whitespace.

\subsection{SVGX-SFT Dataset Analysis}
\label{subsec:dataset_analysis}
Figure~\ref{fig:sft_analysis} provides a comprehensive breakdown of the SVGX-SFT dataset, which is designed for LLM4SVG through supervised fine-tuning. 
Fig.~\ref{fig:sft_analysis}(a): The most common words in the dataset include ``logo'', ``document'', ``button'', and ``download'', suggesting that the dataset primarily consists of UI-related SVG elements. This indicates a strong presence of design-related components, making the dataset particularly useful for training models that need to understand or generate UI elements.
Fig.~\ref{fig:sft_analysis}(b): The dataset underwent a cleaning process, significantly reducing the number of \texttt{<path>} elements and path coordinates. 
This removes redundant or unnecessary elements, improving data quality and reducing complexity.
However, certain elements like \texttt{<rect>}, \texttt{<circle>}, and \texttt{<ellipse>} remained relatively stable, indicating their importance in the dataset.
Fig.~\ref{fig:sft_analysis}(c): The word cloud visualization highlights key terms associated with SVG elements, reinforcing the dominant themes from subfigure (a). Words like ``button'', ``cloud'', ``folder'', and ``icon'' emphasize the dataset's focus on UI-related graphics. The presence of words such as ``geometric'', ``outline'', and ``pattern'' suggests that the dataset also includes a variety of structured vector graphics.
Fig.~\ref{fig:sft_analysis}(d): The pie chart shows the distribution of SVG group counts, with a significant portion (23.1\%) having 5 groups, followed by other group counts such as 4, 6, and 3. This indicates that a considerable number of SVGs in the dataset are structured hierarchically, which is crucial for understanding compositional relationships in vector graphics.

\subsection{Quantitative Evaluation}
\label{sec:qualitative}
Table~\ref{tab:compare_t2v} presents a comparison of our approach with the most prominent text-to-SVG baseline methods across the previously defined dimensions. Our LLM4SVG model achieves the highest performance among LLM-based SVG generation methods. While it may not surpass optimization-based methods in terms of visual quality, it remains competitive, closely matching the performance of other methods across several evaluation metrics. 
Furthermore, our approach only requires model inference, eliminating the need for a time-consuming optimization process, which results in a significantly shorter SVG generation time compared to optimization-based methods.

As shown in table~\ref{tab:compare_llm}, our LLM4SVG outperforms all other LLMs across every aspect. Moreover, due to the limited support for continuous numerical data in most LLMs, SVGs generated by these models often exhibit imprecise coordinates and color representations, typically relying on integers rather than decimals and using basic color names like \texttt{black} or \texttt{blue}, instead of more precise hexadecimal codes. Our approach addresses these shortcomings by incorporating decimal coordinates and hexadecimal color codes for SVG paths, thereby expanding the model’s ability to represent a wider and more accurate range of colors (as illustrated in Fig.~\ref{fig:supp_integer}). Further comparisons with LLM-based methods are provided in Sec.~\ref{sec:supp_compare_LLMs} of Supplementary.

\subsection{Qualitative Evaluation}
\label{sec:quantitative}
Figure~\ref{fig:compare} illustrates the visual quality of SVG generation methods, comparing both optimization-based and LLM-based approaches. It is evident that our method outperforms other LLM-based methods in terms of the completeness of the SVG generation, the selection and placement of primitives, and the semantic richness conveyed by the vector graphics. 
Optimization-based methods~\cite{clipdraw_frans_2022,evolution_tian_2022,diffsketcher_xing_2023,vectorfusion_jain_2023,svgdreamer_xing_2023} use samples from the Latent Diffusion Model as supervision during the SVG generation process. Consequently, these methods normally employ a large number of overlapping and interwoven primitives to closely approximate the realistic samples. 
This often lead to excessive stroke redundancy, and the individual shapes of the primitives may appear irregular when viewed in isolation, making them less practical for use in real-life applications. 

\begin{table}[t]
\centering
\resizebox{1.0\linewidth}{!}{
\begin{tabular}{l|c|c|c|c}
\toprule
Metric/Method & GPT4-o~\cite{GPT4} & Claude-3.5~\cite{claude3.5} & LLM4SVG & Human SVG \\
\midrule
Prompt Alignment & 0.49 & 0.56 & 0.89 & 0.94 \\
Visual Quality & 0.61 & 0.74 & 0.92 & 0.95 \\
Pick Score & 0.57 & 0.69 & 0.88 & 0.92 \\
\bottomrule
\end{tabular}
}
\vspace{-0.5em}
\caption{
\textbf{Results of Human Evaluation.} Human SVG refers to the SVGs designed by human designers in our collected dataset.
} \label{tab:human_eval}
\vspace{-1em}
\end{table}
\noindent\textbf{Human Evaluation.}
To evaluate the visual effects of our generated SVGs compared to other LLMs, we conducted a user study. As Claude 3.5 and GPT-4 are currently recognized as two of the best LLMs, we compared SVGs generated by our LLM4SVG with those produced by these models. In this study, we shuffled the SVGs generated by the different models together with the selected examples from our dataset. Participants were only provided with the text descriptions and the shuffled SVGs. They were then asked to evaluate the SVGs based on prompt alignment, visual quality, and their willingness to use the SVGs in real-world scenarios. 
The results are presented in Table~\ref{tab:human_eval}, where our generated SVGs significantly outperformed other LLM-based methods, and the visual quality of our SVGs is closely comparable to that of SVGs created by humans. Further details of human evaluation are provided in Sec.~\ref{sec:supp_user_study_details} of Supplementary.

\subsection{Ablation Study}
\label{subsec:ablation}
\noindent\textbf{Analysis of the LLM4SVG Architecture.}
The performance of LLM4SVG is significantly influenced by the underlying base model, particularly in how the tokenizer processes continuous numeric tokens. For instance, in the Qwen models~\cite{qwen2_wang_2024,qwen2.5}, all numbers and decimal points are tokenized as a single unit. This approach prevents Qwen from effectively handling continuous numerical data, resulting in poor performance in tasks such as coordinate prediction and hexadecimal color generation. Consequently, the model struggles to produce complete and coherent SVGs.
In contrast, LLMs like GPT-2~\cite{gpt2_radford_2019}, which explicitly enumerate all numbers up to 1,000 as individual tokens, demonstrate a stronger numerical understanding. This design enables them to generate more precise coordinates and color values, leading to more visually coherent and aesthetically refined SVGs.

\noindent\textbf{Analysis of SVG Semantic Tokens.}
To enhance SVG understanding, we expanded the tokenizer with 55 SVG-specific semantic tokens, allowing the model to distinguish SVG tags and attributes from general text. For instance, without this modification, the word ``path'' could ambiguously refer to an SVG tag or a physical pathway. With a dedicated \texttt{<path>} token, the model can now differentiate between the two, improving SVG generation accuracy.

We analyze the impact of these tokens using t-SNE visualization in Sec.~\ref{sec:supp_svg_tokens} of Supplementary. The green dots represent embeddings initialized with descriptions, forming structured clusters for SVG geometry and path command tags. In contrast, the orange crosses, which represent randomly initialized tokens, are scattered, leading to slower convergence. After training, as shown by the blue squares, the embeddings become well-organized, demonstrating that the model effectively learns semantic relationships between SVG elements.

\section{Conclusion \& Discussion}
\label{sec:conclusion}
In this work, we introduced \textit{LLM4SVG}, the first framework designed to support existing LLMs/MLLMs in both understanding and generating SVGs. Our approach allows language models to directly interpret SVG source code and produce high-quality SVGs that demonstrate meaningful complexity and align with human design principles.
Our methodology employs a structured SVG encoding strategy that overcomes the limitations of LLMs treating SVG source code merely as plain text. 
Additionally, we developed the \textit{SVGX-SFT Dataset}, which comprises high-quality SVGs created by human designers and dialogue-based instruction pairs tailored specifically for training LLMs. We anticipate that this dataset will significantly accelerate future research in vector graphics.

\noindent\textbf{Future Work.} LLM4SVG lays the foundation for language models in vector graphics, with several promising directions for future exploration. Enhancing semantic understanding of SVG elements could enable more meaningful editing and generation. Bridging text, raster images, and vector graphics through cross-modal translation would improve creative workflows. Supporting iterative refinement via natural language feedback could make SVG generation more interactive and user-friendly. Additionally, adapting the approach to specific domains like data visualization or UI/UX design could enhance its practical applications.

\noindent\textbf{Acknowledgment.}
This work was supported in part by the National Science and Technology Major Project (No.~2022ZD0117800), the Young Elite Scientists Sponsorship Program by the Chinese Association for Science and Technology (CAST), and the Fundamental Research Funds for the Central Universities. It was also partially funded by the National Natural Science Foundation of China (No.~62461160331, No.~62132001), the Hong Kong Research Grants Council (RGC) General Research Fund (No.~17203023), and the Hong Kong RGC Collaborative Research Fund (No.~C5052-23G). Additionally, this research received support from the National Natural Science Foundation of China/RGC Collaborative Research Scheme (No.~CRS\_HKU703/24), the Huawei-BUAA Joint Lab, and the Hong Kong Jockey Club Charities Trust under Grant No. 2022-0174.

{
    \small
    \bibliographystyle{ieeenat_fullname}
    \bibliography{main}
}

\clearpage
\renewcommand{\thefigure}{S\arabic{figure}}
\setcounter{figure}{0}
\renewcommand{\thetable}{S\arabic{table}}
\setcounter{table}{0}
\maketitlesupplementary

\appendix

\section*{Overview}
\label{sec:overview}
In this supplementary material, we provide additional details and discussions related to our work on \textbf{LLM4SVG}. Specifically, this document covers the following aspects:
\begin{itemize}
\item \textbf{Comparison with Existing LLM-Based Methods} (Sec.~\ref{sec:supp_compare_LLMs}): We demonstrate the advantages of our approach over existing LLM-based methods for SVG generation, particularly in terms of visual appeal and the ability to handle numerical coordinates.
    
\item \textbf{Dataset and Preprocessing Pipeline} (Sec.~\ref{sec:supp_datasets}): We introduce our newly collected dataset and describe a lossless preprocessing pipeline that enhances data quality for training and evaluation.
    
\item \textbf{SVG Semantic Tokens} (Sec.~\ref{sec:supp_svg_tokens}): We provide details on the SVG semantic tokens used in our model, along with an analysis of how different word embedding initialization methods affect model performance.
    
\item \textbf{User Study Details} (Sec.~\ref{sec:supp_user_study_details}): We present additional details about the user study, including the number of participants, their backgrounds, and the methodology used to obtain evaluation metrics.
    
\item \textbf{Primitive Ordering in SVG Generation} (Sec.~\ref{sec:supp_primitive_ordering}): We illustrate how our LLM4SVG model generates SVGs by following a structured primitive ordering strategy. We compare two different generation approaches—a step-by-step composition process and a top-down refinement method—and discuss their implications for SVG interpretability and usability.
    
\item \textbf{Additional SVG Generation Results} (Sec.~\ref{sec:supp_additional_results}): We showcase a diverse set of SVG illustrations generated by our model, demonstrating its ability to produce high-quality, semantically accurate, and stylistically consistent vector graphics across various categories, including objects, animals, food, and abstract symbols.
\end{itemize}

\section{Comparison with LLM-based Methods}
\label{sec:supp_compare_LLMs}
Apart from Figure~\ref{fig:compare} in the manuscript, we present additional qualitative comparisons of our method with existing LLM-based methods in this section, including ChatGPT-3.5~\cite{ChatGPT}, GPT-4o~\cite{GPT4}, GPT-o1-preview~\cite{GPT4}, Claude 3.5-sonnet~\cite{claude3.5}, Gemini 1.5-pro~\cite{geminipro}, Grok 2~\cite{Grok2}, LLama 3.1~\cite{LLaMA_Touvron_2023, llama3_dubey_2024}, Qwen 2.5~\cite{qwen2_wang_2024, qwen2.5}, and Yi~\cite{yi_young_2024}.

\begin{figure*}
    \centering
    \includegraphics[width=\linewidth]{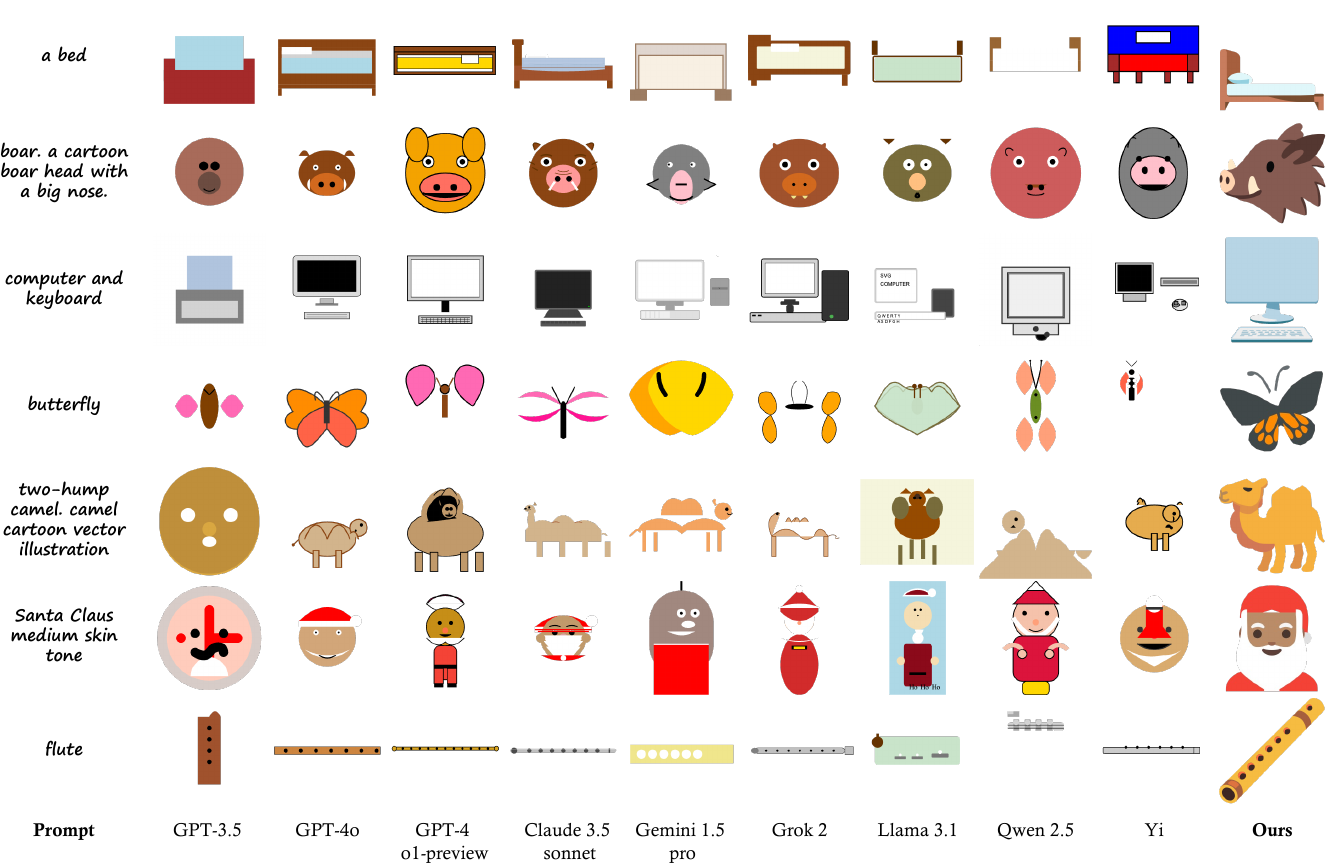}
    \caption{
    \textbf{Qualitative Comparison of LLM4SVG with Existing LLM-based Methods.} Given textual prompts (left), various LLMs generate corresponding vector graphics. The comparison highlights differences in abstraction, structural consistency, and fidelity to the input descriptions. Our LLM4SVG demonstrates improved coherence, accuracy, and stylistic refinement in SVG generation.
    } \label{fig:supp_compare_llm}
\end{figure*}

As illustrated in Fig.~\ref{fig:supp_compare_llm}, it is evident that our proposed LLM4SVG outperforms other methods in terms of overall visual effect and detail expression.
Specifically, most LLMs struggle to generate complete SVG images, for example, \textit{butterfly}, or produce outputs that accurately align with the provided textual descriptions, such as \textit{two-hump camel} illustrated in the last third example.
Some recent LLMs perform relatively better, such as GPT-4o, GPT-o1-preview~\cite{GPT4}, and Claude 3.5-sonnet~\cite{claude3.5}. However, their results are still less satisfactory as the shapes are overly simplistic (e.g., \textit{flute}) and the colors lack harmony (e.g., \textit{a bed}).
In contrast, the results of our LLM4SVG are more complete, with shapes that are more diverse, colors that are more harmonious, and semantics that are more closely aligned with the prompts.

\begin{figure}
    \centering
    \includegraphics[width=1\linewidth]{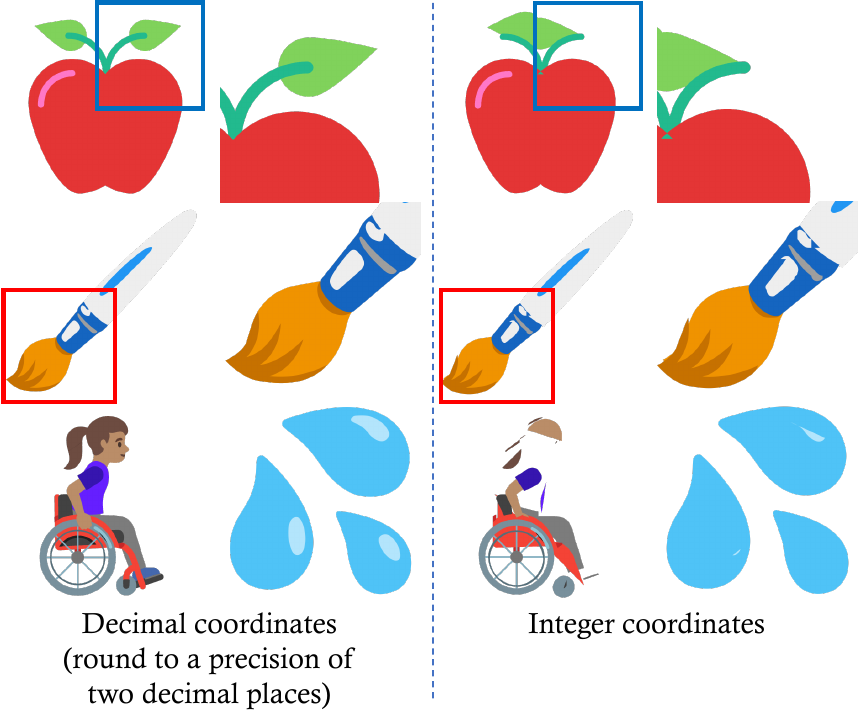}
    \caption{
    \textbf{Visual Comparison between Decimal Coordinates and Integer Coordinates in SVGs.} Only integer coordinates will lead to shape distortions and incompletion.
    } \label{fig:supp_integer}
\end{figure}

\begin{figure*}
    \centering
    \includegraphics[width=\linewidth]{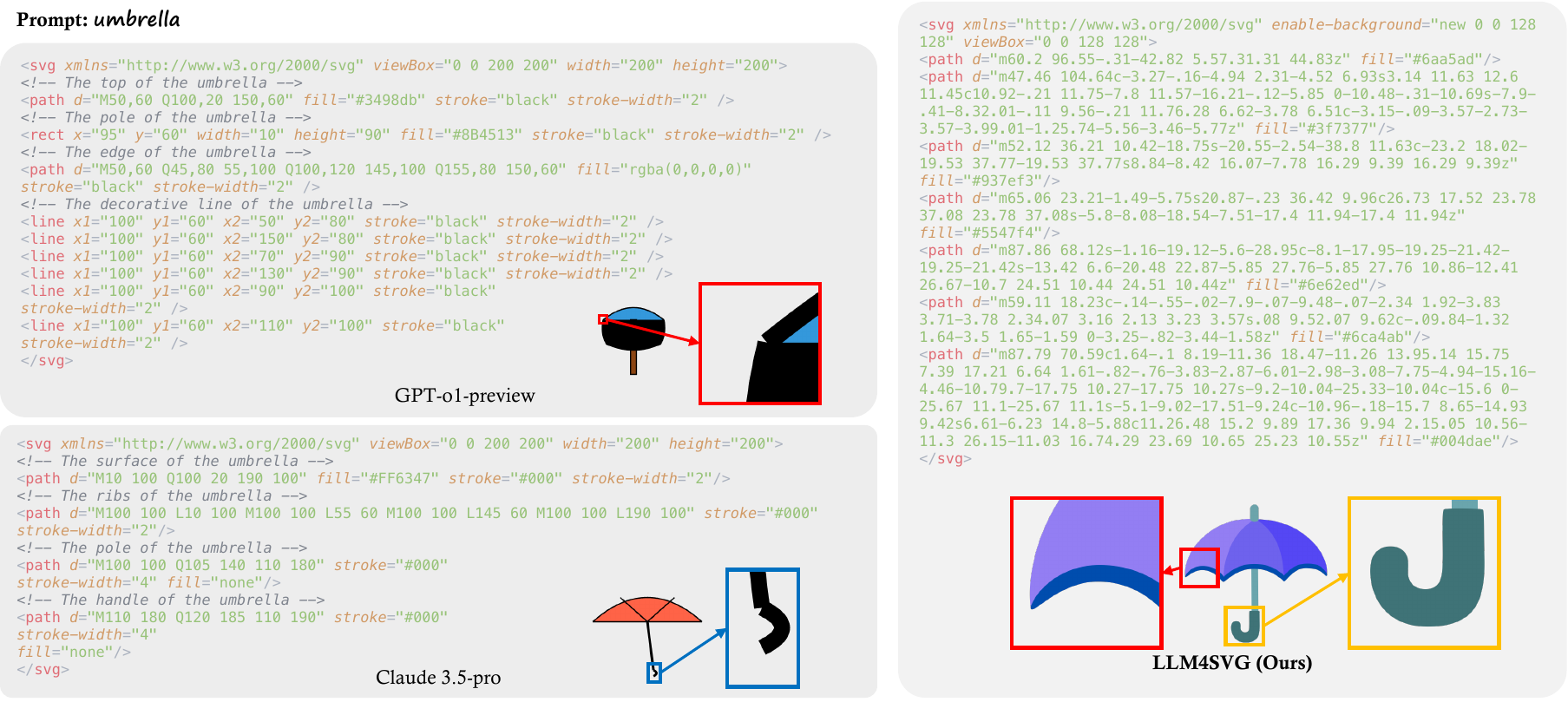}
    \caption{
   \textbf{Example SVG Code Comparison For the Prompt ``umbrella''}.
   The figure contrasts SVG outputs generated by GPT-4o-preview, Claude 3.5-sonnet, and our LLM4SVG. While GPT-4o-preview and Claude 3.5-pro produce basic umbrella-like shapes, their structures contain artifacts or lack refinement. In contrast, LLM4SVG generates a more polished, visually coherent, and structured SVG representation, demonstrating improved detail, smoothness, and geometric accuracy.
    } \label{fig:supp_code}
    \vspace{1em}
\end{figure*}

Additionally, we compare the SVG source codes generated by our method and two of the most recent LLMs, GPT-o1-preview~\cite{GPT4} and Claude 3.5-sonnet~\cite{claude3.5}, to demonstrate the superiority of our method.
As shown in Fig.~\ref{fig:supp_code}, given the prompt ``umbrella'', both GPT-o1-preview~\cite{GPT4} and Claude 3.5-sonnet~\cite{claude3.5} can only predict integer coordinates, whereas our LLM4SVG is capable of generating precise decimal coordinates accurate to two decimal places.
In the context of SVG representation, retaining only integer values can lead to incomplete or distorted SVG shapes, as illustrated in Fig.~\ref{fig:supp_code}.
We present additional examples in Fig.~\ref{fig:supp_integer} to further demonstrate the visual differences between integer and decimal coordinates.

\begin{figure}
    \centering
    \includegraphics[width=1\linewidth]{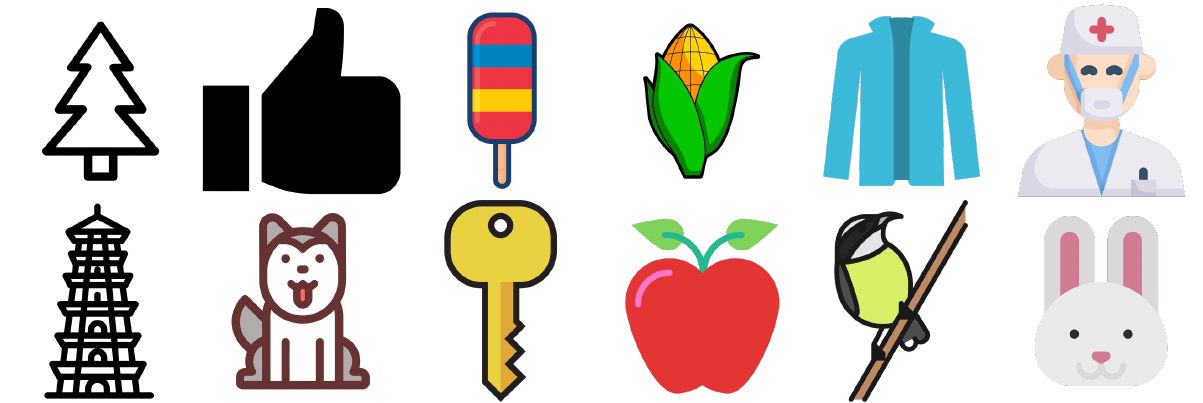}
    \caption{
   \textbf{Illustrative Samples from the SVGX-SFT Dataset}, showcasing its diversity in style, structure, and semantics. The dataset includes a wide range of objects, icons, and illustrations, making it well-suited for training and fine-tuning vector graphic generation models.
    } \label{fig:supp_dataset}
\end{figure}
\begin{figure*}
    \centering
    \includegraphics[width=\linewidth]{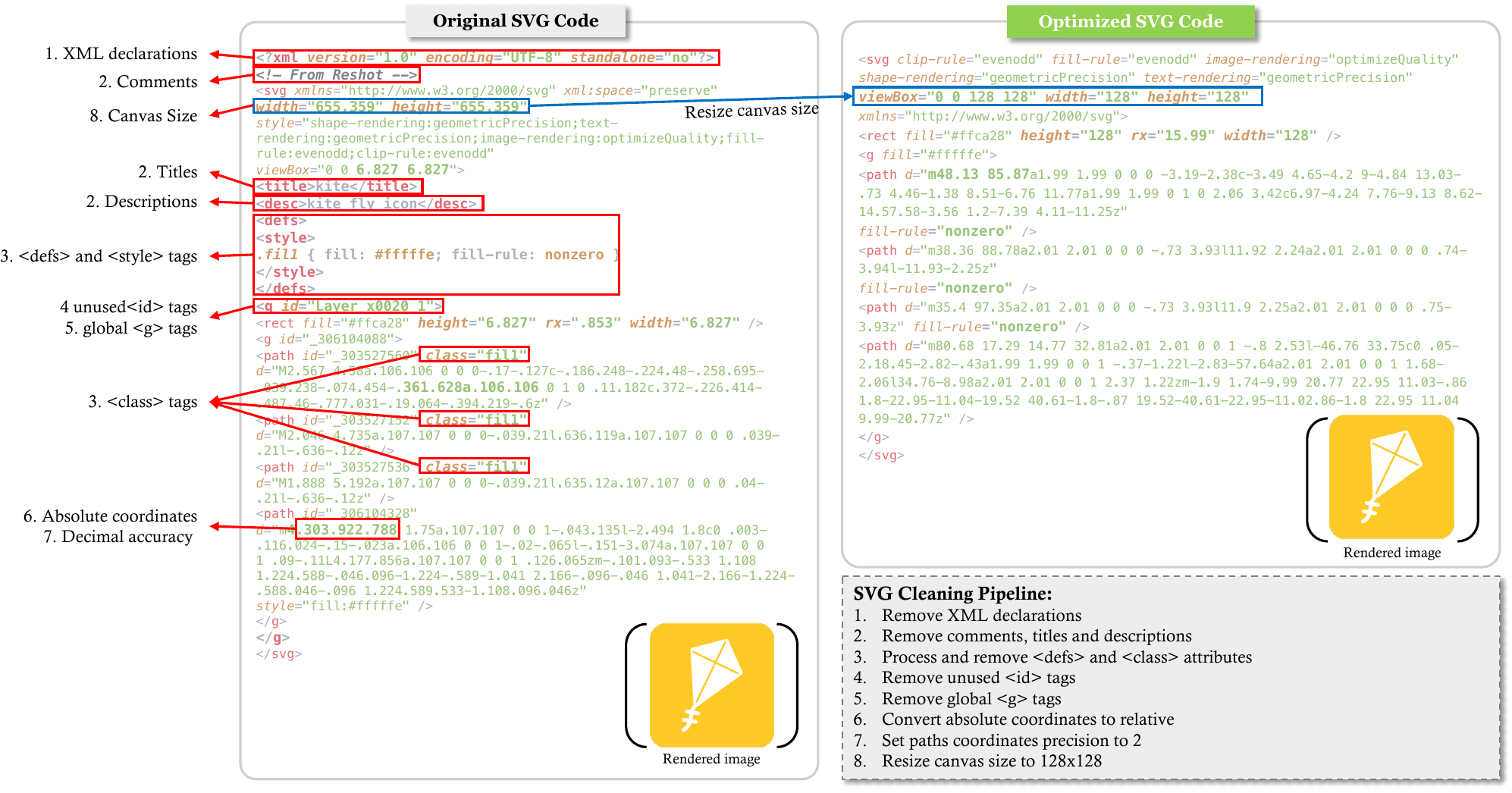}
    \caption{
    \textbf{Illustration of our SVG Processing Pipeline.} The left side shows the original SVG code, which contains redundant elements such as XML declarations, comments, metadata, unused tags, and absolute coordinates. The right side presents the optimized SVG code after applying our cleaning pipeline, which improves efficiency, readability, and scalability by removing unnecessary elements, converting absolute coordinates to relative, and standardizing canvas size. The rendered output remains visually consistent while significantly reducing file complexity.
    }
    \vspace{-1em}
    \label{fig:supp_clean}
\end{figure*}

\section{SVGX-SFT Dataset Details}
\label{sec:supp_datasets}
Figure~\ref{fig:supp_dataset} presents examples from our extensive and diverse SVGX-SFT dataset. This dataset includes primitives of varying complexity, ranging from minimal to highly detailed, while maintaining a rich and harmonious color palette. Additionally, it covers a broad spectrum of subjects, including people, animals, objects, and symbols, making it a comprehensive resource for both SVG generation and understanding tasks.

\noindent\textbf{Preprocessing Pipeline.}  
As discussed in Section~\ref{sec:instruction_data} of our manuscript, a significant portion of an SVG file consists of redundant metadata that does not contribute to its rendered appearance. To improve training efficiency while preserving visual fidelity, we introduce a lossless SVG preprocessing pipeline, as illustrated in Figure~\ref{fig:supp_clean}.

Our pipeline systematically removes unnecessary elements from SVG files, including XML declarations, comments, titles, descriptions, \texttt{<defs>} and \texttt{<class>} tags, as well as global \texttt{<g>} tags. Additionally, we optimize numerical representations by converting absolute coordinates to relative ones and rounding decimal values to a maximum of two decimal places. Furthermore, all SVGs are resized to a standardized $128\times128$ canvas, ensuring consistency across the dataset. These optimizations significantly reduce file size and computational overhead without altering the visual output.

\noindent\textbf{Instruction-Based Dataset Construction.}  
After preprocessing, we structured the dataset to support both SVG generation and understanding tasks. For understanding tasks, each SVG was rasterized into an image, and GPT-4~\cite{GPT4} was used to generate detailed descriptions, which serve as learning targets. For generation tasks, textual prompts were generated using BLIP~\cite{blip_li_2022}, with the corresponding SVG source code serving as the learning content. 

In total, our dataset comprises 580k high-quality SVG-text instruction-compliant samples, providing a robust foundation for training models in structured vector graphic generation and interpretation.

\begin{table*}[!h]
\centering
\resizebox{0.88\linewidth}{!}{
\begin{tabular}{lll}
\toprule
\textbf{Category} & \textbf{Token} & \textbf{Description}  \\
\midrule
\multirow{4}{*}{SVG Container Tags} & \texttt{[<|START\_OF\_SVG|>]} & start of svg \\
 & \texttt{[<|END\_OF\_SVG|>]} & end of svg \\
 & \texttt{[<|start\_of\_g|>]} & start of svg group \\
 & \texttt{[<|end\_of\_g|>]} & end of svg group \\
\midrule
\multirow{8}{*}{SVG Geometry Tags} & \texttt{[<|svg\_path|>]} & svg path element\\
 & \texttt{[<|svg\_circle|>]} & svg circle element\\
 & \texttt{[<|svg\_rect|>]} & svg rectangle element\\
 & \texttt{[<|svg\_ellipse|>]} & svg ellipse element\\
 & \texttt{[<|svg\_polygon|>]} & svg polygon element\\
 & \texttt{[<|svg\_line|>]} & svg line element\\
 & \texttt{[<|svg\_polyline|>]} & svg polyline element\\
 & \texttt{[<|svg\_text|>]} & svg text element\\
\midrule
\multirow{3}{*}{SVG Gradient Tags} & \texttt{[<|svg\_linearGradient|>]} & svg linear gradient element\\
 & \texttt{[<|svg\_radialGradient|>]} & svg radial gradient element\\
 & \texttt{[<|svg\_stop|>]} & svg stop element\\
\midrule
\multirow{10}{*}{Path Commands} & \texttt{[<|moveto|>]} & svg path command, move to \\
 & \texttt{[<|lineto|>]} & svg path command, line to \\
 & \texttt{[<|horizontal\_lineto|>]} & svg path command, horizontal line to \\
 & \texttt{[<|vertical\_lineto|>]} & svg path command, vertical line to \\
 & \texttt{[<|curveto|>]} & svg path command, curve to \\
 & \texttt{[<|smooth\_curveto|>]} & svg path command, smooth curve to \\
 & \texttt{[<|quadratic\_bezier\_curve|>]} & svg path command, quadratic bezier curve \\
 & \texttt{[<|smooth\_quadratic\_bezier\_curveto|>]} & svg path command, smooth quadratic bezier curve \\
 & \texttt{[<|elliptical\_Arc|>]} & svg path command, elliptical arc \\
 & \texttt{[<|close\_the\_path|>]} & svg path command, close the path, close-form\\
\midrule
\multirow{30}{*}{Attribute Tokens} & \texttt{[<|id|>]} & svg element attribute id \\
 & \texttt{[<|d|>]} & svg element attribute define the path \\
 & \texttt{[<|fill|>]} & svg element attribute fill \\
 & \texttt{[<|stroke-width|>]} & svg element attribute stroke-width \\
 & \texttt{[<|stroke-linecap|>]} & svg element attribute stroke-linecap \\
 & \texttt{[<|stroke|>]} & svg element attribute stroke \\
 & \texttt{[<|opacity|>]} & svg element attribute opacity \\
 & \texttt{[<|transform|>]} & svg element attribute transform \\
 & \texttt{[<|gradientTransform|>]} & svg element attribute gradient transform \\
 & \texttt{[<|offset|>]} & svg element attribute offset \\
 & \texttt{[<|width|>]} & svg element attribute width \\
 & \texttt{[<|height|>]} & svg element attribute height \\
 & \texttt{[<|cx|>]} & svg element attribute x coordinate of circle center \\
 & \texttt{[<|cy|>]} & svg element attribute y coordinate of circle center \\
 & \texttt{[<|rx|>]} & svg element attribute x radius of ellipse \\
 & \texttt{[<|ry|>]} & svg element attribute y radius of ellipse \\
 & \texttt{[<|r|>]} & svg element attribute radius of circle \\
 & \texttt{[<|points|>]} & svg element attribute points \\
 & \texttt{[<|x1|>]} & svg element attribute x1 coordinate \\
 & \texttt{[<|y1|>]} & svg element attribute y1 coordinate \\
 & \texttt{[<|x2|>]} & svg element attribute x2 coordinate \\
 & \texttt{[<|y2|>]} & svg element attribute y2 coordinate \\
 & \texttt{[<|x|>]} & svg element attribute x coordinate \\
 & \texttt{[<|y|>]} & svg element attribute y coordinate \\
 & \texttt{[<|fr|>]} & svg element attribute fr \\
 & \texttt{[<|fx|>]} & svg element attribute fx \\
 & \texttt{[<|fy|>]} & svg element attribute fy \\
 & \texttt{[<|href|>]} & svg element attribute href \\
 & \texttt{[<|rotate|>]} & svg element attribute rotate \\
 & \texttt{[<|font-size|>]} & svg element attribute font-size \\
\bottomrule
\end{tabular}
}
\caption{
\textbf{SVG Semantic Tokens Defined by Our LLM4SVG.}
We define 15 tag tokens (including 4 SVG container tags, 8 SVG geometry tags, and 3 SVG gradient tags), 30 attribute tokens, and 10 path command tokens in our LLM4SVG. The ``Token'' field corresponds to the \elem{Token} defined in Table~\ref{tab:instruct_template}. The ``Description'' field is used to initialize the \elem{Token}.
}
\label{tab:supp_svg_tokens}
\end{table*}
\section{Our Proposed SVG Semantic Tokens}
\label{sec:supp_svg_tokens}
For an input SVG $\mathbf{X}_v$, we convert it from raw code into a structured representation.
As shown in Tab.~\ref{tab:supp_svg_tokens}, we present a detailed taxonomy of the SVG semantic tokens employed in LLM4SVG, including 15 tag tokens, 30 attribute tokens and 10 path command tokens.
These SVG tokens are used to replace all tags and attributes in the SVG source code, thus preventing the textual encoding of SVG tags and attributes as regular text. 
This ensures the uniqueness of SVG tags and attributes, and allows for their efficient integration into LLMs in a manner that is consistent with SVG definitions.
The ``Description'' field is utilized to initialize the SVG Tokens based on Equation~\ref{eq:svg_token_init}.

\begin{figure}[t]
\centering
\includegraphics[width=\linewidth]{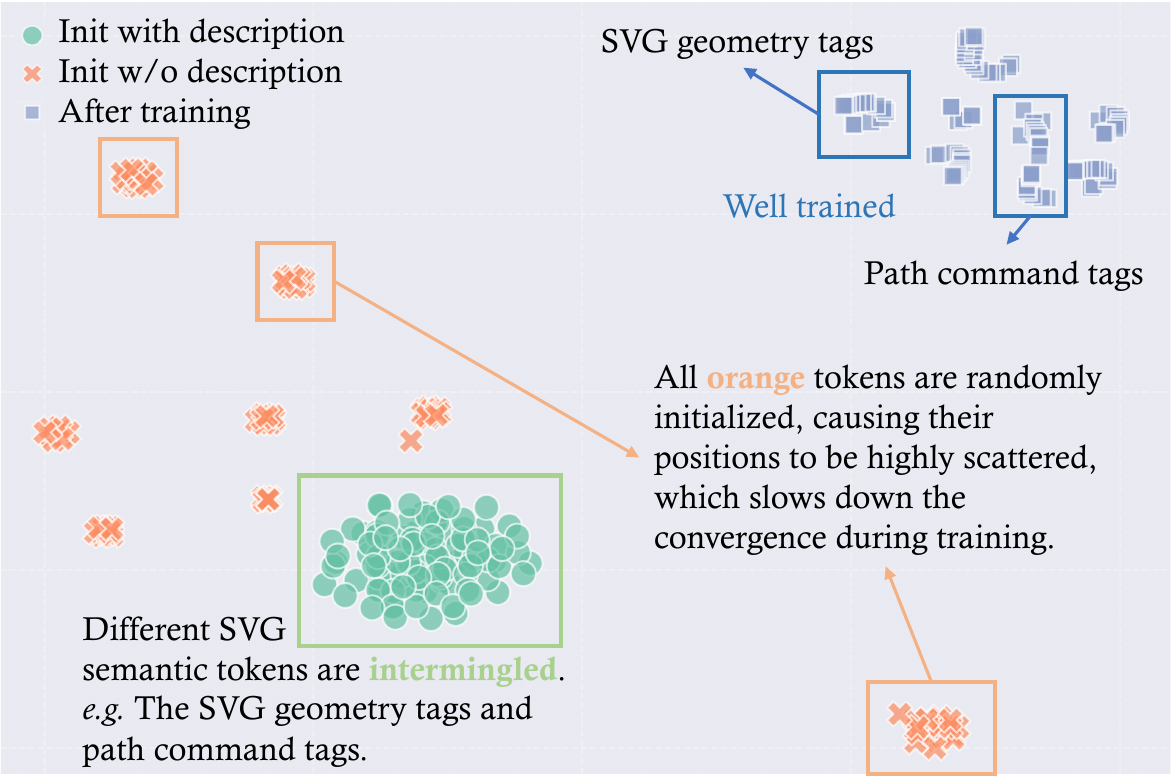}
\caption{
\textbf{t-SNE Visualization of Token Embeddings.}
The green dots represent the SVG token embeddings initialized with descriptions, while the orange crosses indicate those initialized without descriptions. 
The blue squares represent SVG token embeddings after training.
}
\label{fig:tsne}
\vspace{-1em}
\end{figure}
\noindent\textbf{Analysis of Word Embedding Initialization Method.}
As illustrated in Fig.~\ref{fig:tsne}, we used T-SNE~\cite{tsne_van_2008} to map the newly added tokens into a two-dimensional visualization for better demonstration.
In this visualization, the green dots represent tokens that were initialized using the text description of each token. Specifically, we averaged the values obtained from tokenizing these descriptions to initialize the tokens, a process detailed in Sec.~\ref{sec:semantic_token}. This strategy groups the token embeddings into a relatively compact region within the feature space, which helps reduce the difficulty of model training. The orange crosses, on the other hand, indicate tokens that were  initialized without using the text description. These tokens exhibit a more scattered distribution across the feature space, making it challenging for the model to learn the accurate meaning of these tokens.
The blue squares represent the positions of the tokens within the feature space post-training. It is evident that after the training phase, the token embeddings show more meaningful groupings, where tokens with similar semantics are clustered together while maintaining relative proximity to all SVG tokens. Additionally, these tokens remain closely within the overall feature space. This visualization demonstrates the rationale behind our token addition method and the effectiveness of our training approach.

\begin{figure}
    \centering
    \includegraphics[width=1\linewidth]{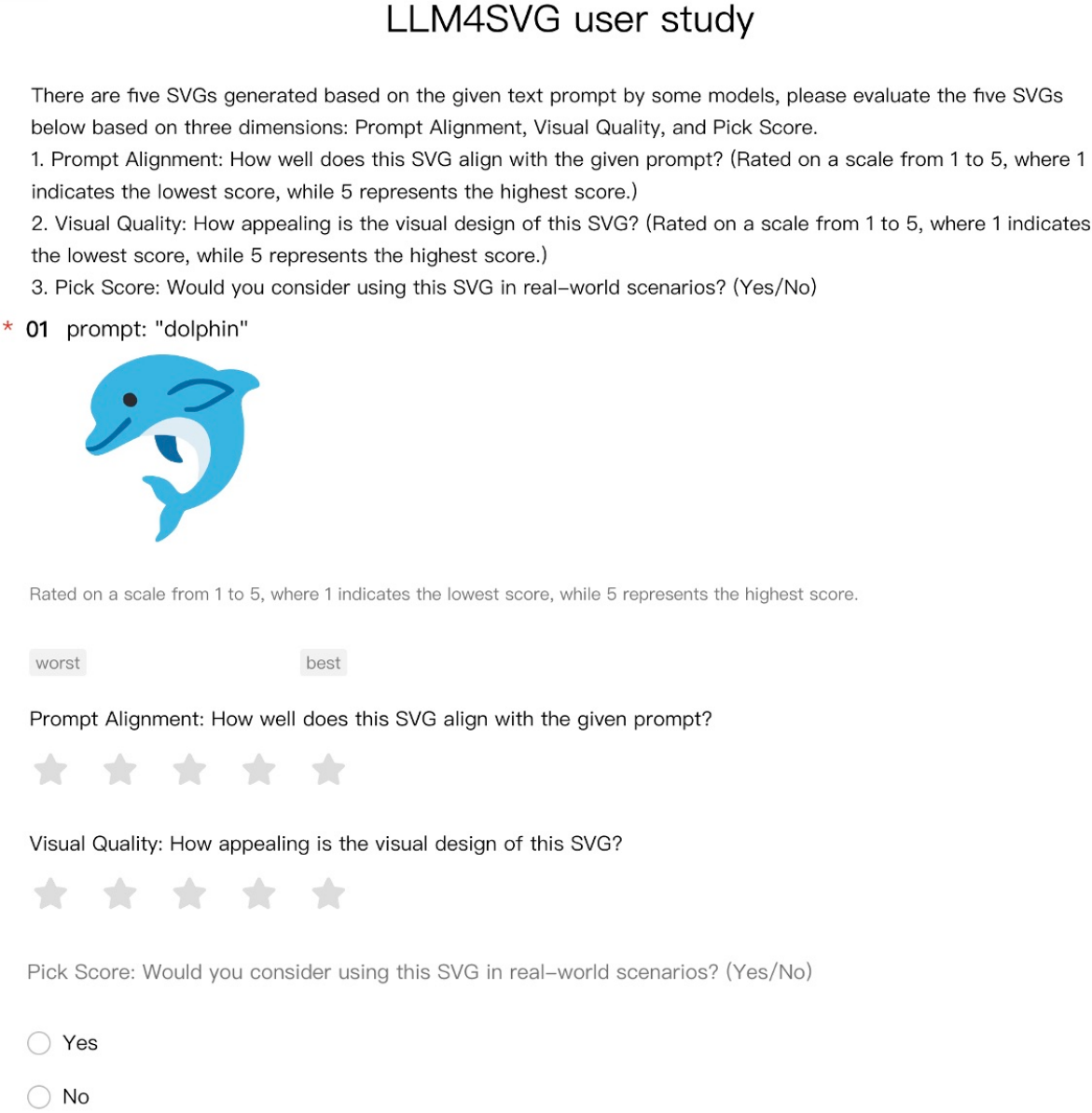}
    \caption{
    \textbf{Screenshot of Our Questionnaire used in the LLM4SVG User Study.}
    Participants evaluate five SVGs generated from a given text prompt based on three key criteria: (1) Prompt Alignment—how well the SVG matches the given prompt, (2) Visual Quality—the aesthetic appeal of the SVG, and (3) Pick Score—whether the participant would consider using the SVG in real-world applications. Ratings are provided on a 5-point scale, with an additional Yes/No selection for practical usability.
    }\label{fig:supp_questionnaire}
    \vspace{-1em}
\end{figure}
\section{More Details about User Study}
\label{sec:supp_user_study_details}
We conducted a user survey to evaluate the effectiveness and practicality of the SVGs generated by our method LLM4SVG and two other popular LLMs, GPT-4o~\cite{GPT4} and Claude-3.5~\cite{claude3.5}.
Specifically, the user study was structured as follows:
\begin{enumerate}
\item \textbf{Data Preparation}: We randomly sampled 17 text descriptions from the evaluation dataset and generated corresponding SVGs using the three models.
Along with the original 17 SVGs from the dataset, this provided a total of 68 SVGs for the user study. 
\item \textbf{Questionnaire Design}: 
Each questionnaire displayed 5 SVGs, randomly sampled from the pool of 68 SVGs. These SVGs were not necessarily generated from the same prompt.

As illustrated in Fig.~\ref{fig:supp_questionnaire}, participants were asked to evaluate each SVG on three aspects:
\begin{itemize}
    \item \textbf{Prompt Alignment}: How well does this SVG align with the given prompt? (Rated on a scale from 1 to 5, where 1 indicates the lowest score, while 5 represents the highest score.)
    \item \textbf{Visual Quality}: How appealing is the visual design of this SVG? (Rated on a scale from 1 to 5, where 1 indicates the lowest score, while 5 represents the highest score.)
    \item \textbf{Pick Score}: Would you consider using this SVG in real-world scenarios? (Yes/No)
\end{itemize}
\item \textbf{Result Calculation}:
This user study involves 37 volunteers from backgrounds in computer science and the arts. 
Each volunteer was required to complete between 1 and 3 questionnaires.
Scores presented in Table 3 of our manuscript were calculated by averaging the ratings for SVGs within the same category. 
For ``Prompt Alignment'' and ``Visual Quality'', the ratings were adjusted by a coefficient $\alpha=0.2$, such that a score of 1 translates to 0.2, and a score of 5 counts to 1.
For ``Pick Score'', a ``Yes'' was scored as 1, while a ``No'' was scored as 0.
\end{enumerate}

\section{Primitive Ordering in SVG Generation}
\label{sec:supp_primitive_ordering}
\begin{figure}
    \centering
    \includegraphics[width=1.0\linewidth]{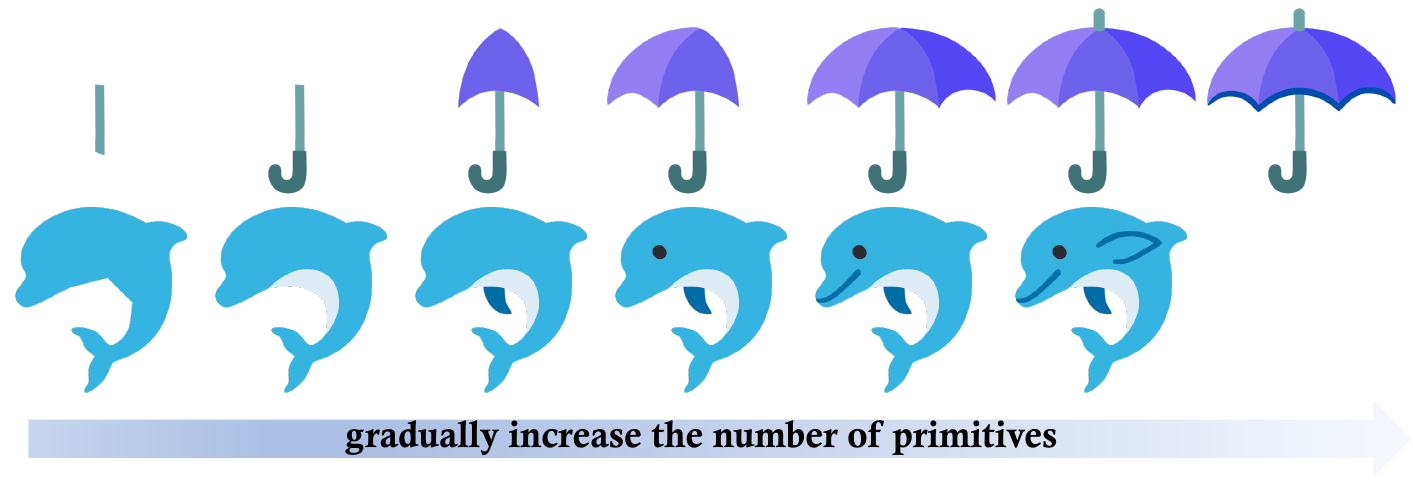}
    \caption{\textbf{Our LLM4SVG model generates SVGs with primitive ordering that aligns with human design principles.}
    The upper example demonstrates a gradual design process, progressing from individual components to the complete design. In contrast, the lower example begins with the overall design before detailing each individual component.}
    \vspace{-1em}
    \label{fig:primitive_ordering}
\end{figure}

Figure~\ref{fig:primitive_ordering} illustrates how our LLM4SVG model generates SVGs by following a primitive ordering strategy that aligns with human design principles. This structured approach ensures that vector graphics are constructed in an intuitive and interpretable manner.

The upper sequence in Figure~\ref{fig:primitive_ordering} demonstrates a step-by-step composition process. The design begins with a few basic primitives, such as lines and simple shapes, and progressively adds more details. This method closely resembles how human designers create illustrations—starting with foundational elements before refining them into a complete object. For example, the umbrella starts with a simple handle, then adds canopy sections until the final structured design emerges. Similarly, the dolphin illustration begins with a basic shape, followed by gradual refinements to enhance realism. This process ensures that each step maintains logical continuity, making the SVG more comprehensible and easier to manipulate.

In contrast, the lower sequence showcases an alternative approach where the entire object is introduced first, followed by successive refinements. This method mimics a top-down design strategy, where an overall form is quickly established before adding intricate details. While this approach can be efficient for certain applications, it lacks the structured progression seen in human sketching workflows, which typically emphasize incremental construction.

By structuring SVG generation in a way that mirrors human cognitive processes, our LLM4SVG model enhances the interpretability and usability of vector graphics. This structured ordering makes the model particularly useful for applications such as educational tools that teach step-by-step drawing, design software that supports intuitive vector editing, and automated graphic generation systems that produce clear and logically constructed icons. The ability to generate illustrations progressively ensures that the output is both aesthetically pleasing and functionally adaptable.

\section{Additional Results Generated by Our LLM4SVG}
\label{sec:supp_additional_results}
\begin{figure*}
    \centering
    \includegraphics[width=\linewidth]{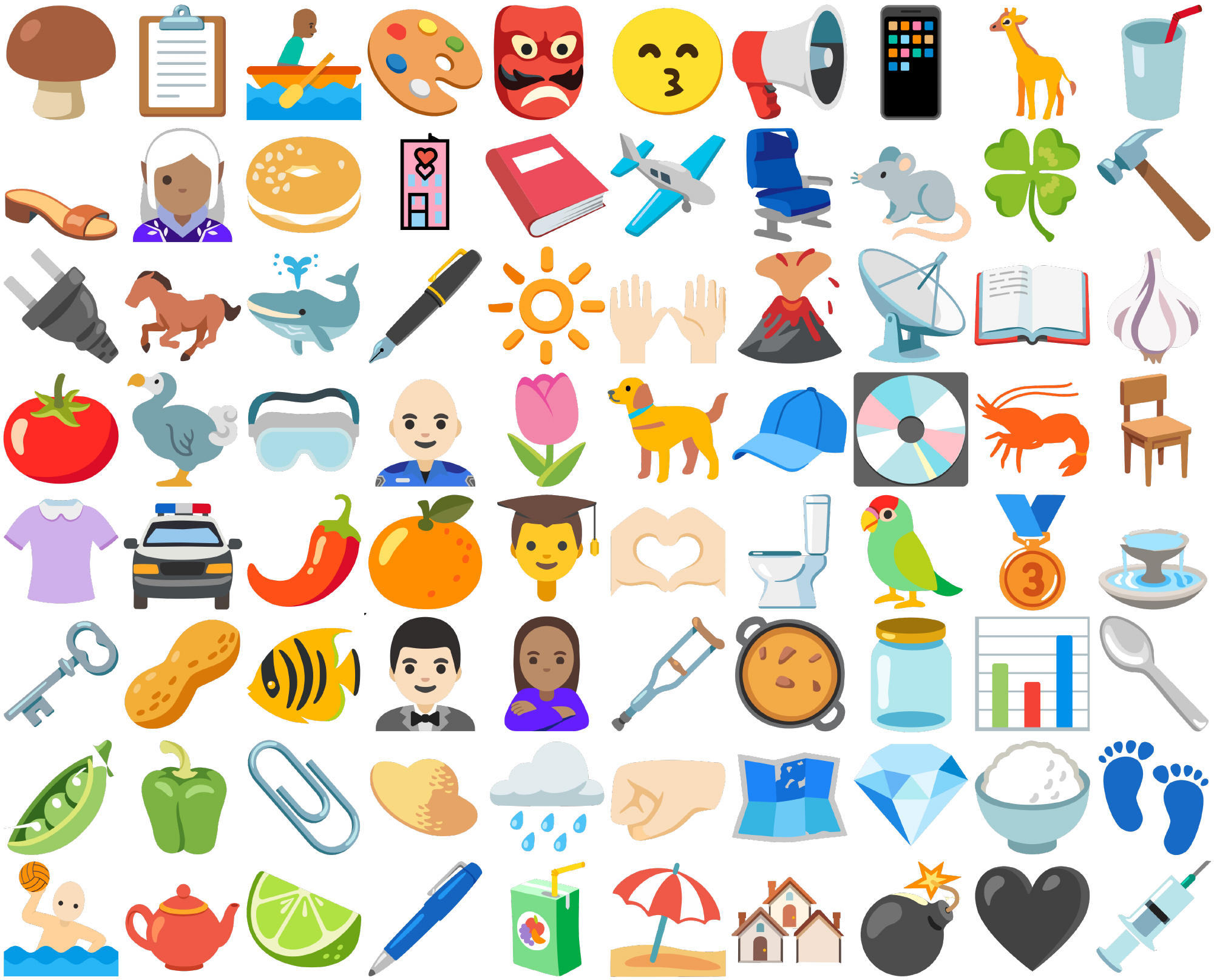}
    \caption{
    \textbf{Additional results generated by our LLM4SVG model.} This collection showcases a diverse range of high-quality SVG illustrations covering various categories, including objects, animals, food, people, symbols, and abstract concepts. The results demonstrate the model's ability to produce visually appealing, semantically accurate, and stylistically consistent vector graphics.
    }
    \label{fig:supp_results}
\end{figure*}

Figure~\ref{fig:supp_results} showcases a diverse collection of SVG illustrations generated by our LLM4SVG model. These results demonstrate the model’s capability to produce high-quality, semantically accurate, and visually appealing vector graphics across a wide range of categories.
The generated SVGs span various domains, including:
\begin{itemize}
    \item \textbf{Objects:} Everyday items such as a pen, paperclip, key, and teapot.
    \item \textbf{Animals:} Illustrations of a dolphin, parrot, horse, fish, and giraffe.
    \item \textbf{Food:} Fruits, vegetables, and prepared dishes, including a tomato, orange, lime, and a rice bowl.
    \item \textbf{People and Expressions:} Human figures representing different professions, emotions, and activities.
    \item \textbf{Symbols and Abstract Concepts:} Medals, graphs, weather symbols, and a heart icon.
\end{itemize}

These results highlight the effectiveness of our model in generating stylistically consistent and visually coherent vector graphics. The illustrations maintain a balanced level of abstraction while preserving key details, making them suitable for real-world applications such as digital icons, UI elements, and educational materials.



\end{document}